\documentclass[runningheads]{llncs}
\pdfoutput=1
\usepackage[dvipsnames]{xcolor}
\usepackage{graphicx}
\usepackage{wrapfig}
\usepackage{comment}
\usepackage{amsmath,amssymb} 
\usepackage{color}
\usepackage[stable]{footmisc}

\usepackage[pagebackref=true,breaklinks=true,letterpaper=true,colorlinks,bookmarks=false]{hyperref}

\usepackage{arydshln}
\usepackage{booktabs}
\usepackage{enumitem}
\usepackage{balance}
\usepackage{combelow}
\usepackage{subfigure}
\usepackage{tabularx}
\usepackage{cite}
\usepackage{pifont}

\usepackage{multicol}
\usepackage{multirow}
\usepackage{makecell}
\usepackage{algorithm}
\usepackage[noend]{algpseudocode}
\usepackage{algorithmicx,eqparbox,array}

\usepackage[dvipsnames]{xcolor}
\usepackage{colortbl}

\definecolor{Gray}{gray}{0.9}
\DeclareMathOperator*{\argmin}{\arg\!\min}

\newcommand{\cmark}{\ding{51}}%
\newcommand{\xmark}{\ding{55}}%

\makeatletter
\DeclareRobustCommand\onedot{\futurelet\@let@token\@onedot}
\def\@onedot{\ifx\@let@token.\else.\null\fi\xspace}

\makeatother

\newcommand\blfootnote[1]{%
  \begingroup
  \renewcommand\thefootnote{}\footnote{#1}%
  \addtocounter{footnote}{-1}%
  \endgroup
}

\makeatletter
\def\BState{\State\hskip-\ALG@thistlm}
\makeatother

\definecolor{Gray}{gray}{0.9}

\begin{document}
\pagestyle{headings}
\mainmatter
\def\ECCVSubNumber{1652}  

\title{Learning Architectures for Binary Networks\\
\vspace{0.5em}\small \url{https://github.com/gistvision/bnas}\vspace{-1em}} 

\titlerunning{Learning Architectures for Binary Networks}
%
\author{
Dahyun Kim$^*$\inst{1}
\and
Kunal Pratap Singh$^*$\inst{2}
\and
Jonghyun Choi\inst{1}
}
\authorrunning{D. Kim et al.}
%
\institute{GIST, South Korea \and
IIT Roorkee, India\\
\email{killawhale@gm.gist.ac.kr, ksingh@ee.iitr.ac.in,jhc@gist.ac.kr}
}

\maketitle

\begin{abstract}
Backbone architectures of most binary networks are well-known floating point architectures such as the ResNet family.
Questioning that the architectures designed for floating point networks would not be the best for binary networks, we propose to search architectures for binary networks (BNAS)
by defining a new search space for binary architectures and a novel search objective.
Specifically, based on the cell based search method, we define the new search space of binary layer types, design a new cell template, and rediscover the utility of and propose to use the \emph{Zeroise} layer instead of using it as a placeholder. 
The novel search objective \textit{diversifies early search} to learn better performing binary architectures. 
We show that our proposed method searches architectures with stable training curves despite the quantization error inherent in binary networks.
Quantitative analyses demonstrate that our searched architectures outperform the architectures used in state-of-the-art binary networks and outperform or perform \emph{on par} with state-of-the-art binary networks that employ various techniques other than architectural changes.
\blfootnote{\hspace{-1.8em}$^*$ indicates equal contribution. This work is done while KPS is at GIST for internship.}
\end{abstract}

\section{Introduction}
\label{sec:intro}

Increasing demand for deploying high performance visual recognition systems encourages research on efficient neural networks.
Approaches include pruning\cite{han2015learning}, efficient architecture design\cite{zhang2018shufflenet,2016_SqueezeNet,howard2017mobilenets}, low-rank decomposition\cite{jaderberg2014speeding}, network quantization\cite{courbariaux2015binaryconnect,Rastegari2016XNORNetIC,li2016ternary} and knowledge distillation\cite{hinton2015distilling,tan2018learning}. 
Particularly, network quantization, especially binary or 1-bit CNNs, are known to provide extreme computational and memory savings. The computationally expensive floating point convolutions are replaced with computationally efficient XNOR and bit-count operations, which significantly speeds up inference\cite{Rastegari2016XNORNetIC}.
Hence, binary networks are incomparable with efficient floating point networks due to the extreme computational and memory savings.

Current binary networks, however, use architectures designed for floating point weights and activations\cite{Rastegari2016XNORNetIC,liu2018bi,Liu2019CirculantBC, shen2019searching}. 
We hypothesize that the backbone architectures used in current binary networks may not be optimal for binary parameters as they were designed for floating point ones. 
Instead, we may learn better binary network architectures by exploring the space of binary networks. 
\begin{figure}[!t]
\centering
\includegraphics[width=0.68\columnwidth]{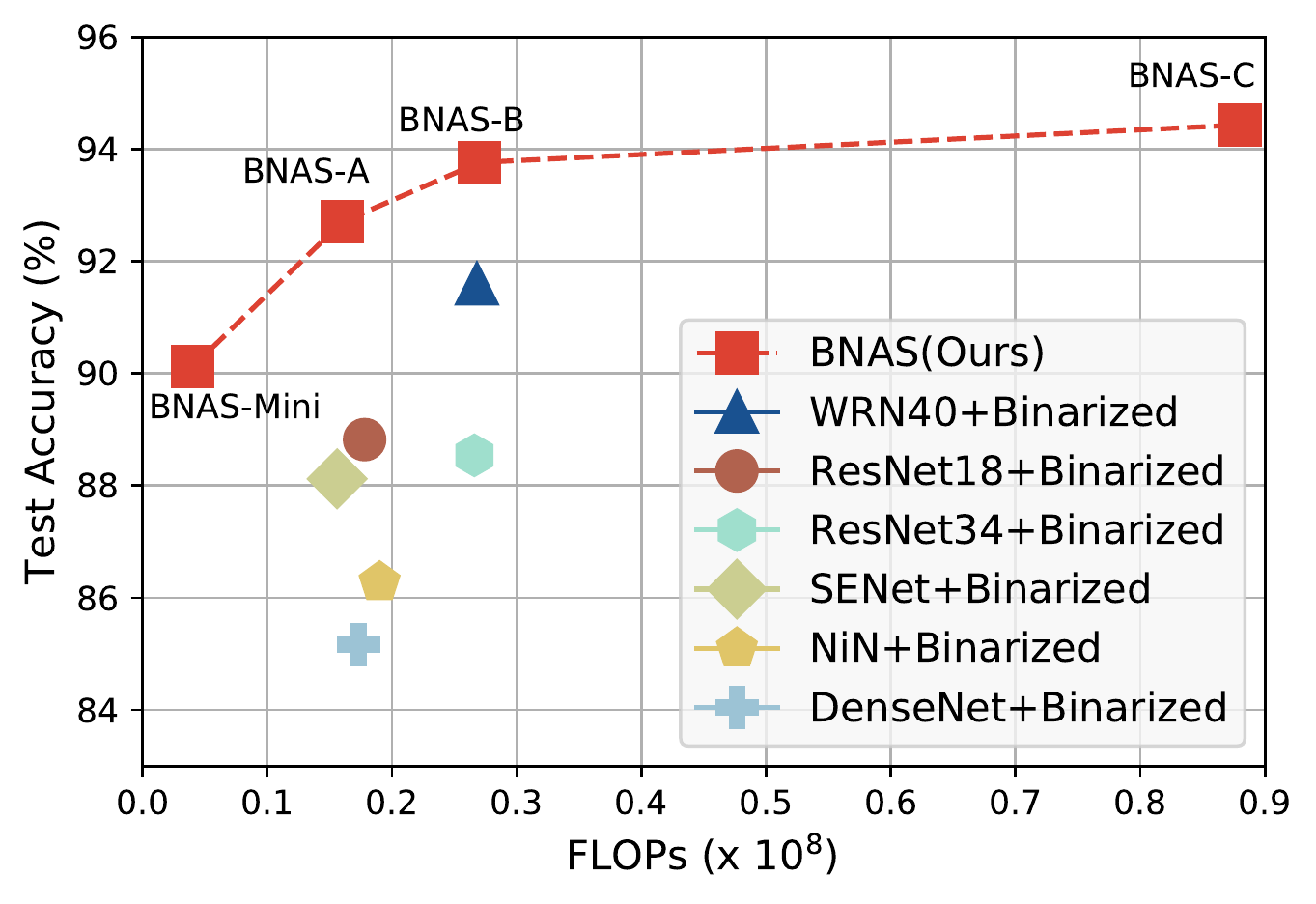}\\
\vspace{-1.5em}
\caption{\textbf{Test accuracy (\%) vs. FLOPs on CIFAR10 for various backbone architectures binarized using the XNOR-Net binarization scheme \cite{Rastegari2016XNORNetIC}}. Our searched architectures outperform the binarized floating point architectures. Note that our BNAS-Mini, which has much less FLOPs, outperforms all other binary networks except the one based on WideResNet40 (WRN40)}
\label{fig:teaser}
\vspace{-2em}
\end{figure}

To discover better performing binary networks, we first apply one of the widely used binarization schemes~\cite{Rastegari2016XNORNetIC} to the searched architectures from floating point NAS which use cell based search and gradient based search algorithms~\cite{liu2018darts,xie2018snas,dong2019search}. We then train the resulting binary networks on CIFAR10.
Disappointingly, the binarized searched architectures do not perform well (Sec.~\ref{sec:binarizing_fpnas}).
We hypothesize two reasons for the failure of binarized searched floating point architectures.
First, the search space used in the floating point NAS is not necessarily the best one for binary networks. 
For example, separable convolutions will have large quantization error when binarized, since nested convolutions increase quantization error (Sec.~\ref{sec:sep_conv}).
Additionally, we discover that the \textit{Zeroise} layer, which was only used as a placeholder in floating point NAS, improves the accuracy of binary networks when kept in the final architecture (Sec.~\ref{sec:zeroise}). 
Second, the cell template used for floating point cell based NAS methods is not well suited for the binary domain because of unstable gradients due to quantization error (Sec.~\ref{sec:cell_design}).

Based on the above hypotheses and empirical observations, we formulate a cell based search space explicitly defined for binary networks and further propose a novel search objective with the diversity regularizer.
The proposed regularizer encourages exploration of diverse layer types in the early stages of search, which is particularly useful for discovering better binary architectures.
We call this method as Binary Network Architecture Search or \emph{BNAS}.
We show that the new search space and the diversity regularizer in BNAS helps in searching better performing binary architectures (Sec.~\ref{sec:exp}).

Given the same binarization scheme, we compare our searched architectures to several handcrafted architectures including the ones shown in the Fig.~\ref{fig:teaser}.
Our searched architectures clearly outperforms the architectures used in the state-of-the-art binary networks, indicating the prowess of our search method in discovering better architectures for binary networks.

\vspace{0.5em}
\noindent We summarize our contributions as follows:
\vspace{-0.5em}
\renewcommand{\labelitemi}{$\bullet$}
\begin{itemize}[leftmargin=0.8\parindent]
    \item We propose the first architecture search method for binary networks. The searched architectures are adjustable to various computational budgets (in FLOPs) and outperform backbone architectures used in state-of-the-art binary networks on both CIFAR10 and ImageNet dataset.
    
    \item We define a new search space for binary networks that is more robust to quantization error; a new cell template and a new set of layers. 
    
    \item We propose a new search objective aimed to diversify early stages of search and demonstrate its contribution in discovering better performing binary networks.
\end{itemize}

\vspace{-1em}
\section{Related Work}
\label{sec:related}

\subsection{Binary Neural Networks}

There have been numerous proposals to improve the accuracy of binary (1-bit) precision CNNs whose weights and activations are all binary valued.
We categorize them into binarization schemes, architectural modifications and training methods.

\vspace{0.5em}\noindent\textbf{Binarization Schemes.}
As a pioneering work, \cite{courbariaux2015binaryconnect} proposed to use the sign function to binarize the weights and achieved compelling accuracy on CIFAR10.
\cite{courbariaux2016binarized} binarized the weights and the activations by the sign function and use the straight through estimator (STE) to estimate the gradient.
\cite{Rastegari2016XNORNetIC} proposed XNOR-Net which uses the sign function with a scaling factor to binarize the weights and the activations. 
They showed impressive performance on a large scale dataset (ImageNet ILSVRC 2012) and that the computationally expensive floating point convolution operations can be replaced by highly efficient XNOR and bit counting operations.
Many following works including recent ones~\cite{liu2018bi,Liu2019CirculantBC} use the binarization scheme of XNOR-Net as do we.
\cite{lin2017towards} approximated both weights and activations as a weighted sum of multiple binary filters to improve performance. 
Very recently, new binarization schemes have been proposed \cite{gu2019projection,bulat2019xnor}. 
\cite{gu2019projection} uses projection convolutional layers while \cite{bulat2019xnor} improves upon the analytically calculated scaling factor in XNOR-Net.

These different binarization schemes do not modify the backbone architecture while we focus on finding better backbone architectures given a binarization scheme. 
A newer binarization scheme can be incorporated into our search framework but that was not the focus of this work.

\vspace{0.5em}\noindent\textbf{Architectural Advances.}
It has been shown that appropriate modifications to the backbone architecture can result in great improvements in accuracy \cite{Rastegari2016XNORNetIC,Liu2019CirculantBC,liu2018bi}.
\cite{Rastegari2016XNORNetIC} proposed XNOR-Net which shows that changing the order of batch normalization (BN) and the sign function is crucial for the performance of binary networks.
\cite{liu2018bi} connected the input floating point activations of consecutive blocks through identity connections before the sign function. 
They aimed to improve the representational capacity for binary networks by adding the floating point activation of the current block to the consequent block.  
They also introduced a better approximation of the gradient of the sign function for back-propagation. 
\cite{Liu2019CirculantBC} used circulant binary convolutions to enhance the representational capabilities of binary networks.
\cite{phan2019mobinet} proposed a modified version of separable convolutions to binarize the MobileNetV1 architecture.
However, we observe that the modified separable convolution modules do not generalize to architectures other than MobileNet.
Most recently, \cite{shen2019searching} use evolutionary algorithms to change the number of channels for each convolution layer of a binarized ResNet backbone.
However, their method does not change the backbone architecture (using ResNet18) but find its hyper-parameters, while we focus on finding better backbone architectures. 
Further, they trade more computation cost for better performance, reducing their inference speed up ($\sim 2.7\times $) to be far smaller than other binary networks ($\sim 10\times$).
These methods do not alter the connectivity or the topology of the network while we search for entirely new network architectures.

\vspace{0.5em}\noindent\textbf{Training Methods.}
There have been a number of methods proposed for training binary networks.
\cite{Zhuang_2018_CVPR} showed that quantized networks, when trained progressively from higher to lower bit-width, do not get trapped in a local minimum. 
\cite{gu2019bayesian} proposed a training method for binary networks using two new losses; Bayesian kernel loss and Bayesian feature loss.
Recently, \cite{Kim2020BinaryDuo} proposed to pretrain the network with ternary activation which are later decoupled to binary activations for fine-tuning.
The training methods can be used in our searched networks as well, but we focus on the architectural advances.

\vspace{-0.5em}
\subsection{Efficient Neural Architecture Search}
We search architectures for binary networks by adopting ideas from neural architecture search (NAS) methods for floating point networks \cite{Zoph_2018_CVPR,liu2018darts,xie2018snas,leiclr2017,pmlr-v80-pham18a}.
To reduce the severe computation cost of NAS methods, there are numerous proposals focused on accelerating the NAS algorithms \cite{liu2018darts,dong2019search,xie2018snas,cai2018proxylessnas,pmlr-v80-bender18a,luo_nips_2018,Liu_2019_CVPR,wu2019fbnet,zhou2018resource,Liu_2018_ECCV,dong2018dpp,liu2018hierarchical,zhang2019graph,xie2019exploring,li2019random,pmlr-v80-pham18a}.
We categorize these attempts
into cell based search and gradient based search algorithms.

\vspace{0.5em}\noindent\textbf{Cell Based Search.}
Pioneered by \cite{Zoph_2018_CVPR}, many NAS methods \cite{xie2018snas,dong2019search,pmlr-v80-bender18a,luo_nips_2018,Liu_2019_CVPR,liu2018darts,wu2019fbnet,zhou2018resource,Liu_2018_ECCV,dong2018dpp,liu2018hierarchical,zhang2019graph,xie2019exploring,li2019random} have used the cell based search, where the objective of the NAS algorithm is to search for a cell, which will then be stacked to form the final network.
The cell based search reduces the search space drastically from the entire network to a cell, significantly reducing the computational cost. 
Additionally, the searched cell can be stacked any number of times given the computational budget. 
Although the scalability of the searched cells to higher computational cost is a non-trivial problem~\cite{Tan2019EfficientNetRM}, it is not crucial to our work because binary networks focus more on smaller computational budgets.

\vspace{0.5em}\noindent\textbf{Gradient Based Search Algorithms.}
In order to accelerate the search, methods including \cite{liu2018darts,xie2018snas,wu2019fbnet,dong2019search} relax the discrete sampling of child architectures to be differentiable so that the gradient descent algorithm can be used.
The relaxation involves taking a weighted sum of several layer types during the search to approximate a single layer type in the final architecture.
\cite{liu2018darts} uses softmax of learnable parameters as the weights, while other methods \cite{xie2018snas,wu2019fbnet,dong2019search} use the Gumbel-softmax \cite{gumbel} instead, both of which allow seamless back-propagation by gradient descent.
Coupled with the use of the cell based search, certain work has been able to drastically reduce the search complexity~\cite{dong2019search}.

We make use of both the cell based search and gradient based search algorithms but propose a novel search space along with a modified cell template and a new regularized search objective to search binary networks.

\vspace{-0.5em}
\section{Binarizing Searched Architectures by NAS}
\label{sec:binarizing_fpnas}

\begin{wrapfigure}{R}{0.5\textwidth}
\vspace{-3.5em}
\begin{center}
\includegraphics[width=0.24\columnwidth]{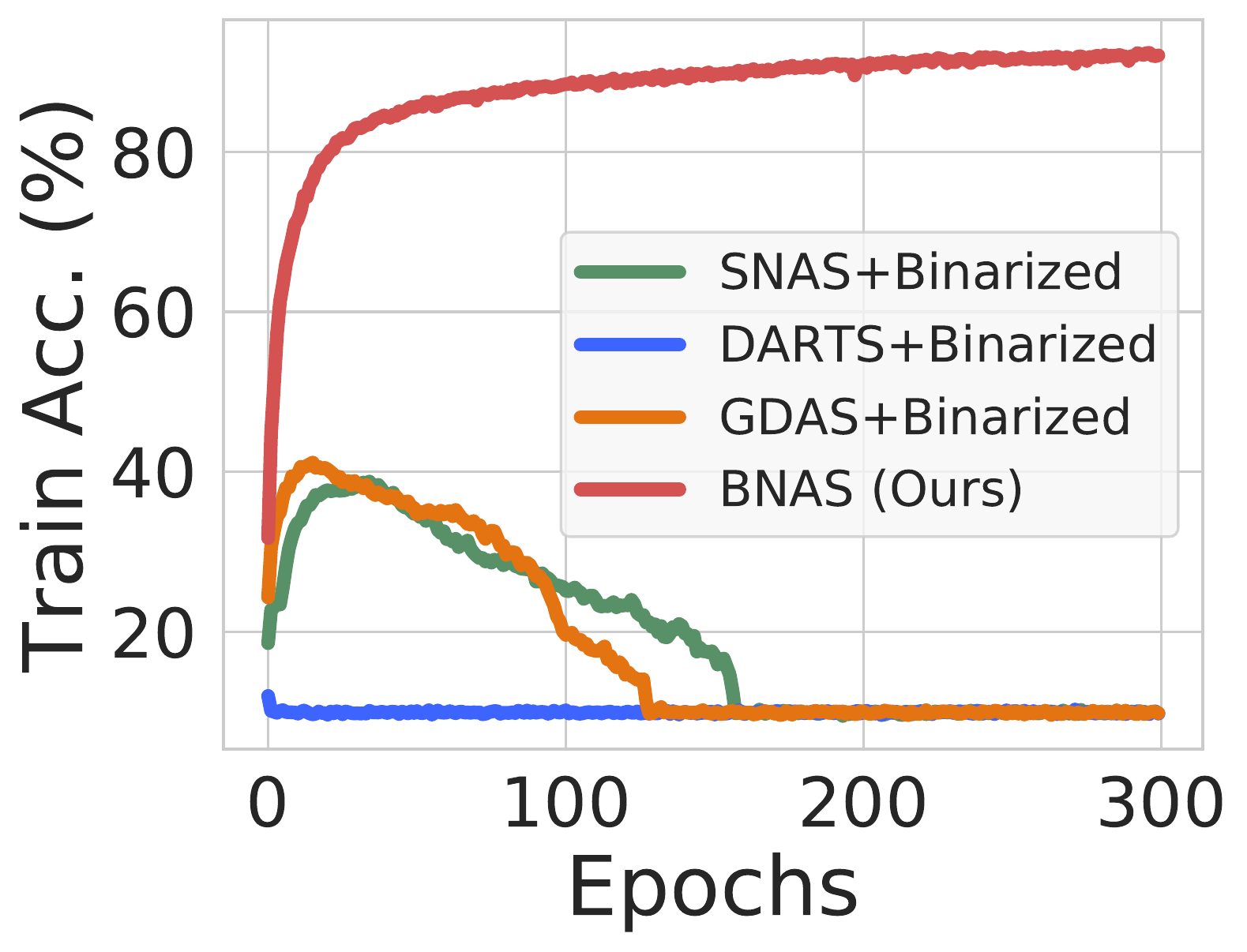}
\includegraphics[width=0.24\columnwidth]{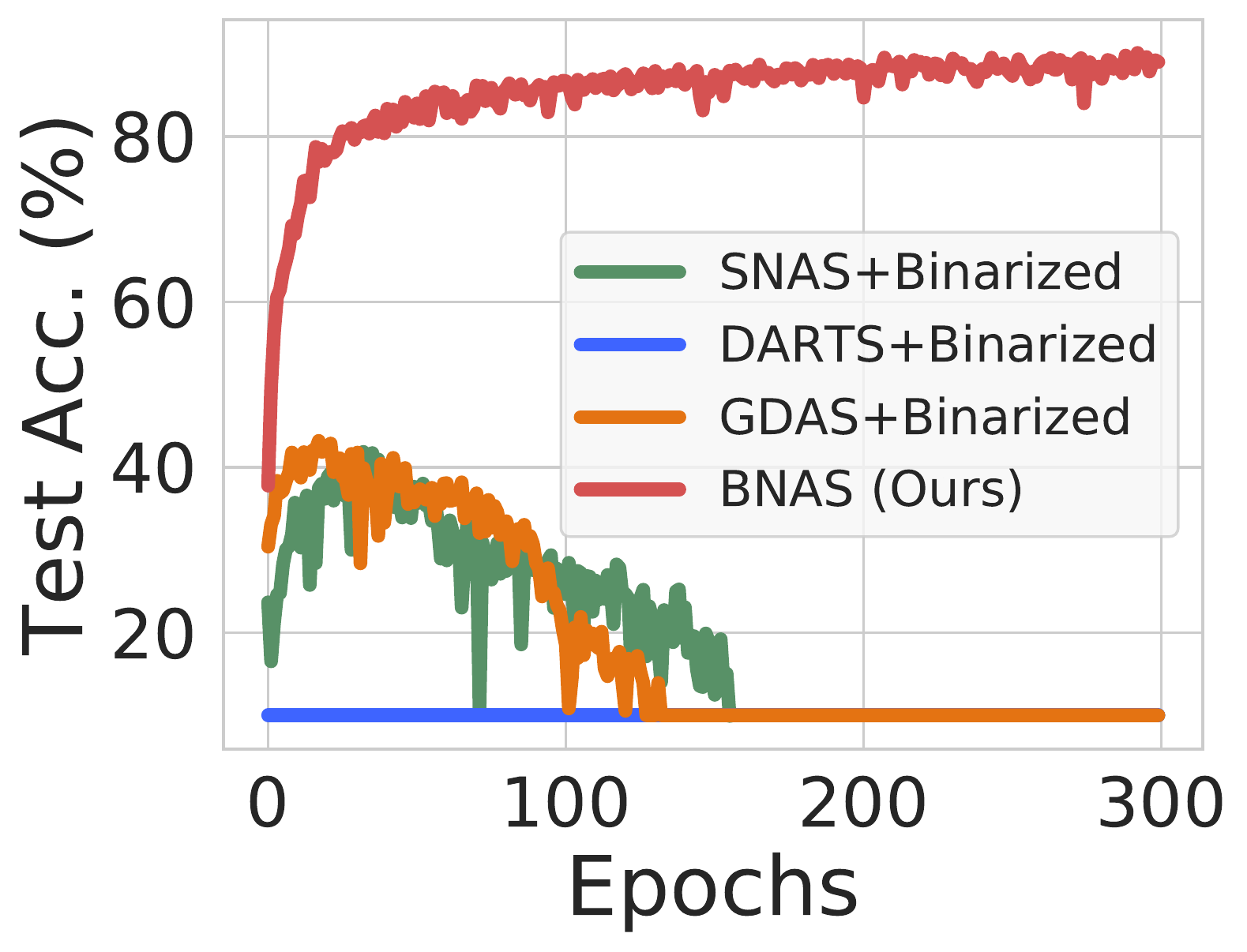}\\
\end{center}
\vspace{-15pt}
\caption{\textbf{Train (left) and test (right) accuracy of binarized searched architectures on CIFAR10}. The XNOR-Net's binarization scheme and architectural modifications are applied in all cases. Contrasting to our BNAS, the binarized searched architectures fail to train well}
\vspace{-20pt}
\label{fig:search_constraint}
\end{wrapfigure}

It is well known that architecture search results in better performing architecture than the hand-crafted ones. 
To obtain better binary networks, we first binarize the searched architectures by cell based gradient search methods.
Specifically, we apply the binarization scheme of XNOR-Net along with their architectural modifications~\cite{Rastegari2016XNORNetIC} to architectures searched by DARTS, SNAS and GDAS. 
We show the learning curves of the binarized searched floating point architectures on CIFAR10 dataset in Fig.~\ref{fig:search_constraint}. 

Disappointingly, GDAS and SNAS reach around $40\%$ test accuracy and quickly plummet while DARTS did not train at all.
This implies that floating point NAS methods are not trivially extended to search binary networks.
We investigate the failure modes in training and find two issues; 
1) the search space is not well suited for binary networks, \textit{e.g}, using separable convolutions accumulates the quantization error repetitively and 
2) the cell template does not propagate the gradients properly, due to quantization error.
To search binary networks, the search space and the cell template should be redesigned to be robust to quantization error.

\vspace{-0.5em}
\section{Approach}
\label{sec:approach} 

To search binary networks, we first write the problem of cell-based architecture search in general as:
\begin{equation}
    \alpha^{*} = \displaystyle \argmin_{\alpha \in A(\mathcal{S}, T)} \mathcal{L}_S(D; \theta_{\alpha}),
\end{equation}
where $A$ is the feasible set of final architectures, $\mathcal{S}$ is the search space (a set of layer types to be searched), $T$ is the cell template which is used to create valid networks from the chosen layer types, $L_S$ is the search objective, $D$ is the dataset, $\theta_{\alpha}$ is the parameters of the searched architecture $\alpha$ which contain both architecture parameters (used in the continuous relaxation~\cite{liu2018darts}, Eq.~\ref{eq:diversity}) and the network weights (the learnable parameters of the layer types, Eq.~\ref{eq:diversity}), and $\alpha^*$ is the searched final architecture.
Following \cite{liu2018darts}, we solve the minimization problem using SGD.

Based on the observation in Sec.~\ref{sec:binarizing_fpnas}, we propose a new search space ($\mathcal{S}_B$), cell template ($T_B$) and  a new search objective $\widetilde{\mathcal{L}}_S$ for binary networks which have binary weights and activations.
The new search space and the cell template are more robust to quantization error and the new search objective $\widetilde{\mathcal{L}}_S$ promotes diverse search which is important when searching binary networks (Sec. \ref{sec:diver_search}).
The problem of architecture search for binary network $\alpha^{*}_B$ can be rewritten as:
\begin{equation}
    \alpha_B^{*} = \displaystyle \argmin_{\alpha_B \in A_B(\mathcal{S}_B, T_B)} \widetilde{\mathcal{L}}_S(D; \theta_{\alpha_B}),
\end{equation}
where $A_B$ is the feasible set of binary network architectures and $\theta_{\alpha_B}$ is parameters of the binary networks.
We detail each proposal in the following subsections.

\vspace{-1em}
\subsection{Search Space for Binary Networks ($\mathcal{S}_B$)}
\label{sec:layer_types}

Unlike the search space used in floating point NAS, the search space used for binary networks should be robust to quantization error.
Starting from the search space popularly used in floating point NAS~\cite{liu2018darts,xie2018snas,dong2019search,Zoph_2018_CVPR}, we investigate the robustness of various convolutional layers to quantization error and selectively define the space for the binary networks.
Note that the quantization error depends on the binarization scheme and we use the scheme proposed in~\cite{Rastegari2016XNORNetIC}.

\vspace{-1em}
\subsubsection{Convolutions and Dilated Convolutions.}
\label{sec:conv}

To investigate the convolutional layers' resilience to quantization error, we review the binarization scheme we use~\cite{Rastegari2016XNORNetIC}.
Let ${\bf W}$ be the weights of a floating point convolution layer with dimension $c \cdot w \cdot h$ (number of channels, width and height of an input) and ${\bf A}$ be an input activation.
The floating point convolution can be approximated by binary parameters, ${\bf B}$, and the binary input activation, ${\bf I}$ as:
\begin{equation}
    {\bf W * A} \approx \beta {\bf K} \odot ({\bf B * I}),
\label{eq:bin_conv}
\end{equation}
where $*$ denotes the convolution operation, $\odot$ is the Hadamard product (element wise multiplication), ${\bf B} = sign({\bf W})$, ${\bf I} = sign({\bf A})$, $\beta = \frac{1}{n} \| {\bf W}\|_1$ with $n = c \cdot w \cdot h$, ${\bf K} = {\bf D * k}$, ${\bf D} = \frac{\sum |A_{i,:,:}|}{c}$ and ${\bf k}_{ij} =\frac{1}{w \cdot h}$ $\forall ij$.
Dilated convolutions are identical to convolutions in terms of quantization error.

Since both convolutions and dilated convolutions show tolerable quantization error in binary networks~\cite{Rastegari2016XNORNetIC} (and our empirical study in Table~\ref{table:search_space}), we include the standard convolutions and dilated convolutions in our search space.

\vspace{-1em}
\subsubsection{Separable Convolutions.}
\label{sec:sep_conv}

Separable convolutions~\cite{sifre2014rigid} have been widely used to construct efficient network architectures for floating point networks~\cite{howard2017mobilenets} in both hand-crafted and NAS methods.
Unlike floating point networks, we argue that the separable convolution is not suitable for binary networks due to large quantization error.
It uses nested convolutions to approximate a single convolution for computational efficiency. 
The nested convolution are approximated to binary convolutions as:
\begin{equation}
    Sep({\bf W * A}) \approx \beta_2 ({\bf B}_2 * {\bf A}_2)
                     \approx \beta_1 \beta_2 ({\bf B}_2 * ({\bf K}_1 \odot ({\bf B}_1 * {\bf I}_1))), \\
\label{eq:sep_conv}
\end{equation}
where $Sep({\bf W * A})$ denotes the separable convolution, ${\bf B}_1$ and ${\bf B}_2$ are the binary weights for the first and second convolution operation in the separable convolution layer, ${\bf I}_1 = sign({\bf A})$, ${\bf A}_2 = \beta_1 {\bf K}_1 \odot ({\bf B}_1 * {\bf I}_1)$
and $\beta_1, \beta_2, {\bf K}_1$ are the scaling factors for their respective binary weights and activations.
Since every scaling factor induces quantization error, the nested convolutions in separable convolutions will result in more quantization error.

To empirically investigate how the quantization error affects training for different convolutional layer types, we construct small networks formed by repeating each kind of convolutional layers three times, followed by three fully connected layers. 
We train these networks on CIFAR10 in floating point and binary domain and summarize the results in Table~\ref{table:search_space}. 

\begin{table}[t!]
\centering
\caption{\textbf{Test Accuracy (\%) of a small CNN composed of each layer type only, in floating point (FP Acc.) and in binary domain (Bin. Acc) on CIFAR10.} \textit{Conv}, \textit{Dil. Conv} and \textit{Sep. Conv} refer to the convolutions, dilated convolutions and separable convolutions, respectively. Separable convolutions show a drastically low performance compared on the binary domain}
\vspace{-.5em}
\begin{tabular}{ccccccc}
\toprule
Layer Type                   & \multicolumn{2}{c}{Conv} & \multicolumn{2}{c}{Dil. Conv} & \multicolumn{2}{c}{Sep. Conv}  \\ 

Kernel Size                  & $3\times3$         & $5\times5$         & $3\times3$            & $5\times5$           & $3\times3$            & $5\times5$                       \\ \cmidrule(lr){1-1} \cmidrule(lr){2-3} \cmidrule(lr){4-5} \cmidrule(lr){6-7}
FP Acc. (\%)               & $61.78$       & $60.14$       & $56.97$          & $55.17$         & $56.38$          & $57.00$                  \\ 
\rowcolor{Gray}
Bin. Acc. (\%)         & $46.15$       & $42.53$       & $41.02$          & $37.68$         & $10.00$          & $10.00$                  \\ 
\bottomrule
\end{tabular}
\label{table:search_space}
\vspace{-1em}
\end{table}

When binarized, both convolution and dilated convolution layers show only a reasonable drop in accuracy, while the separable convolution layers show performance equivalent to random guessing ($10\%$ for CIFAR10).
The observations in Table~\ref{table:search_space} imply that the accumulated quantization error by the nested convolutions fails binary networks in training.
This also partly explains why the binarized architecture searched by DARTS in Fig. \ref{fig:search_constraint} does not train as it selects a large number of separable convolutions.

\vspace{-1em}
\subsubsection{Zeroise.}
\label{sec:zeroise}

\begin{wrapfigure}{R}{0.50\textwidth}
\vspace{-4em}
\begin{center}
\includegraphics[width=0.98\linewidth]{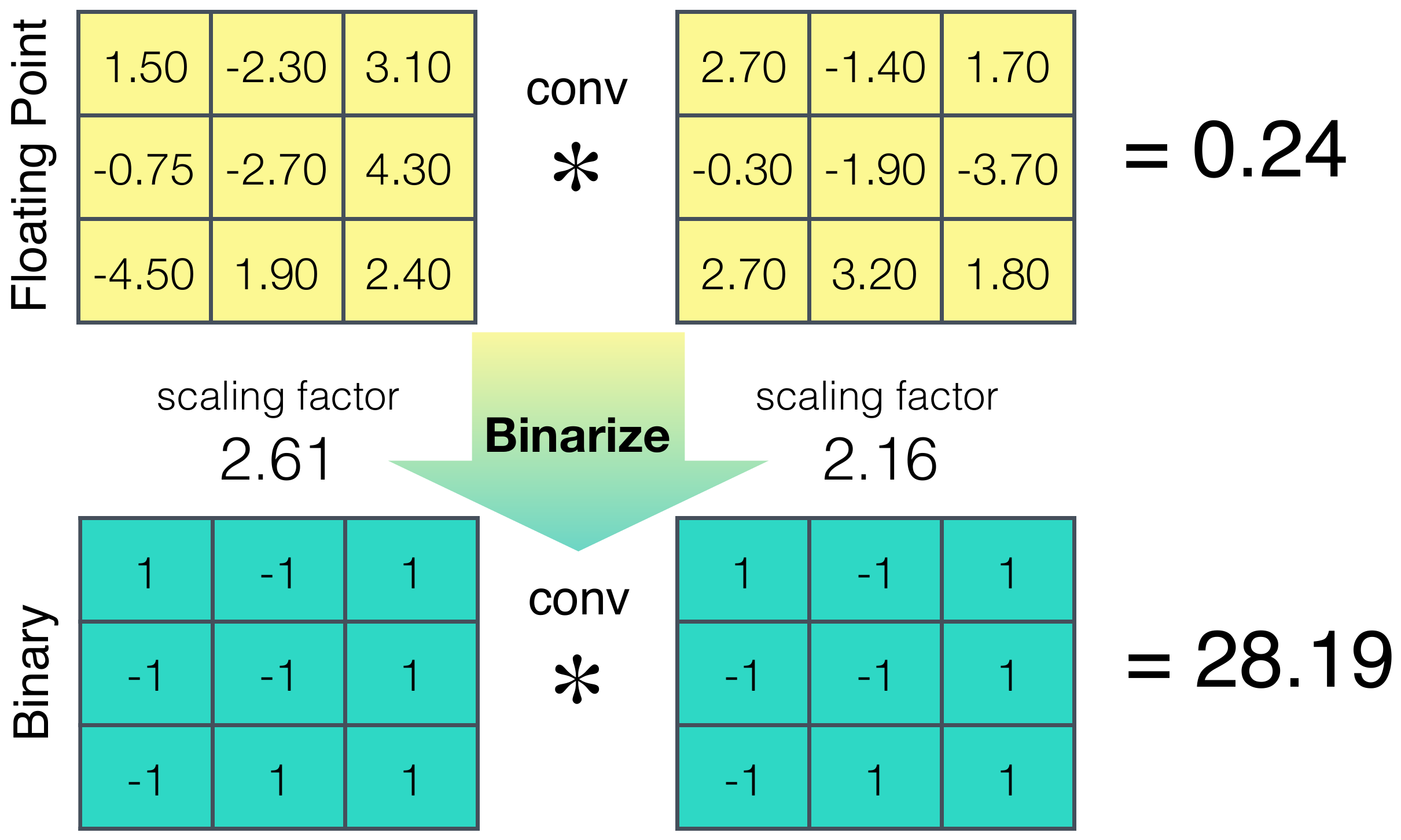}
\end{center}
\vspace{-2em}
\caption{\textbf{An example when the \textit{Zeroise} layer is beneficial for binary networks.} 
Since the floating point convolution is close to zero but the binarized convolution is far greater than 0, if the search selects the \textit{Zeroise} layer instead of the convolution layer, the quantization error reduces significantly}
\vspace{-3em}
\label{fig:zeroise}
\end{wrapfigure}

The \textit{Zeroise} layer outputs all zeros irrespective of the input\cite{liu2018darts}.
It was originally proposed to model the lack of connections.
Further, in the authors' implementation of \cite{liu2018darts}\footnote{\href{https://github.com/quark0/darts}{https://github.com/quark0/darts}.}, the final architecture excludes the \textit{Zeroise} layers and replaces it with the second best layer type, even if the search picks the \textit{Zeroise} layers.
Thus, the \textit{Zeroise} layers are not being used as they were originally proposed but simply used as a placeholder for a different and sub-optimal layer type.
Such replacement of layer types effectively removes all architectures that have \textit{Zeroise} layers from the feasible set of final architectures.

\begin{table}[t!]
\centering
\caption{\textbf{DARTS and BNAS with and without the \textit{Zeroise} layers in the final architecture on CIFAR10.} \textit{Zeroise Layer} indicates whether the \textit{Zeroise} layers were kept ({\color{ForestGreen}\cmark}) or not ({\color{red}\xmark}). The test accuracy of DARTS drops by $3.02\%$ when you include the \textit{Zeroise} layers and the train accuracy drops by $63.54\%$ and the training stagnates. In constrast, the \textit{Zeroise} layers improves BNAS in both train and test accuracy, implying better training without overfitting}
\vspace{-1em}
\begin{tabular}{ccccccc}
\toprule
Precision       & \multicolumn{3}{c}{Floating Point (DARTS)} & \multicolumn{3}{c}{Binary (BNAS)} \\ 
Zeroise Layer       & \color{red}\xmark & \color{ForestGreen}\cmark & Gain &\color{red}\xmark & \color{ForestGreen}\cmark & Gain\\ \cmidrule(lr){1-1} \cmidrule(lr){2-4} \cmidrule(lr){5-7}
Train Acc. (\%) &    $99.18$    &  $35.64$ &{\color{red}-$63.54\%$}     & $93.41$ & $97.46$ &{\color{ForestGreen}+$4.05\%$}  \\ 
Test Acc. (\%)   &     $97.45$    &   $94.43$ &{\color{red}-$3.02\%$}  & $89.47$ & $92.70$ &{\color{ForestGreen}+$3.23\%$}  \\ 
\bottomrule
\end{tabular}
\vspace{-1em}
\label{table:darts_zeroise}
\end{table}

In contrast, we use the \textit{Zeroise} layer for \emph{reducing the quantization error} and are the first to keep it in the \emph{final} architectures instead of using it as a placeholder for other layer types.
As a result, our feasible set is different from that of \cite{liu2018darts} not only in terms of precision (binary), but also in terms of the network topology it contains.

As the exclusion of the \textit{Zeroise} layers is not discussed in \cite{liu2018darts}, we compare the accuracy with and without the \textit{Zeroise} layer for DARTS in the DARTS column of Table~\ref{table:darts_zeroise} and empirically verify that the \textit{Zeroise} layer is not particularly useful for floating point networks.
However, we observe that the \textit{Zeroise} layer improve the accuracy by a meaningful margin in binary networks as shown in the table. 
We argue that the \textit{Zeroise} layer can reduce quantization error in binary networks as an example in Fig.\ref{fig:zeroise}.
Including the \textit{Zeroise} layer in the final architecture is particularly beneficial when the situation similar to Fig. \ref{fig:zeroise} happens frequently as the quantization error reduction is significant.
But the degree of benefit may differ from dataset to dataset.
As the dataset used for search may differ from the dataset used to train and evaluate the searched architecture, we propose to tune the probability of including the \textit{Zeroise} layer.
Specifically, we propose a generalized layer selection criterion to adjust the probability of including the \textit{Zeroise} layer by a transferability hyper-parameter $\gamma$ as:
\begin{equation}
    p^{*} = \max \left[ \frac{p_{z}}{\gamma}, p_{op_1},...,p_{op_n} \right],
    \label{eq:zeroise}
\end{equation}
where $p_z$ is the architecture parameter corresponding to the \textit{Zeroise} layer and $p_{op_i}$ are the architecture parameters corresponding to the $i^{\text{th}}$ layer other than \textit{Zeroise}.
Larger $\gamma$ encourages to pick the \textit{Zeroise} layer only if it is substantially better than the other layers.

With the separable convolutions and the \textit{Zeroise} layer type considered, we summarize the defined search space for BNAS ($\mathcal{S}_B$) in Table~\ref{table:layer_types}.

\subsection{Cell Template for Binary Networks ($T_B$)}
\label{sec:cell_design}

\begin{figure}[t!]
\centering
    \includegraphics[width=0.42\textwidth]{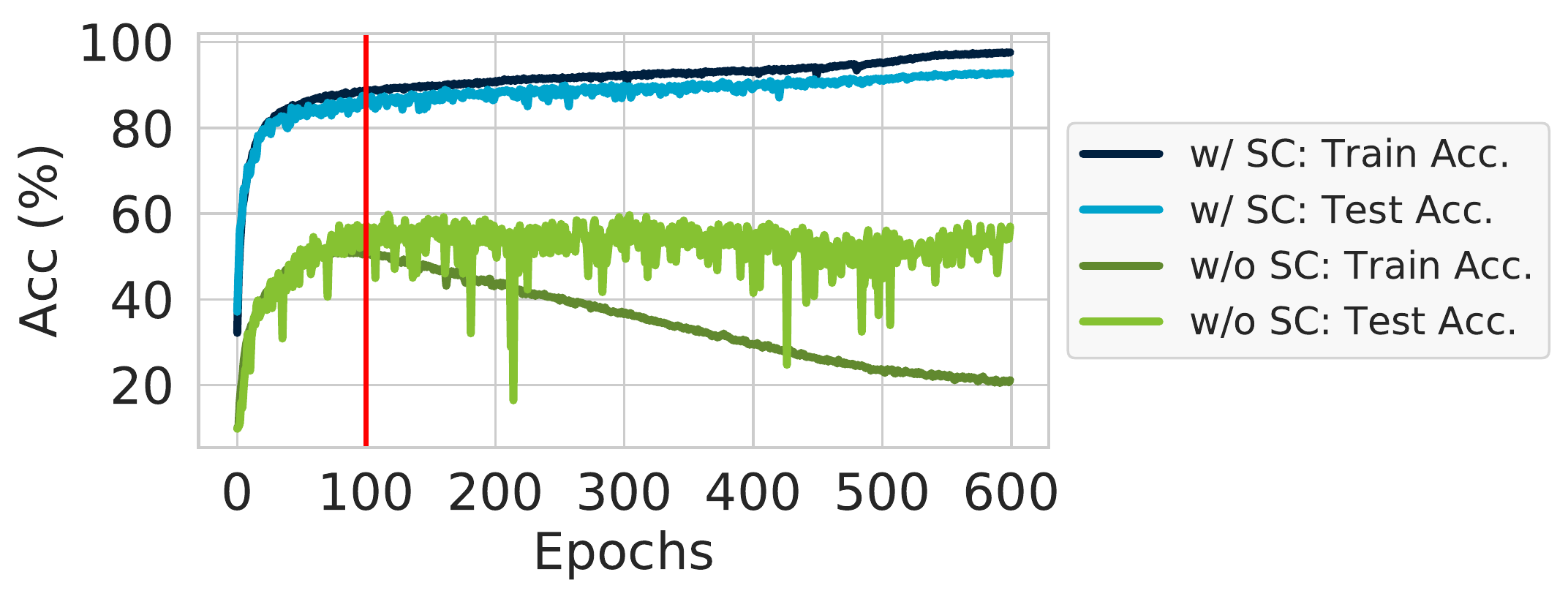}
    \includegraphics[width=0.28\textwidth]{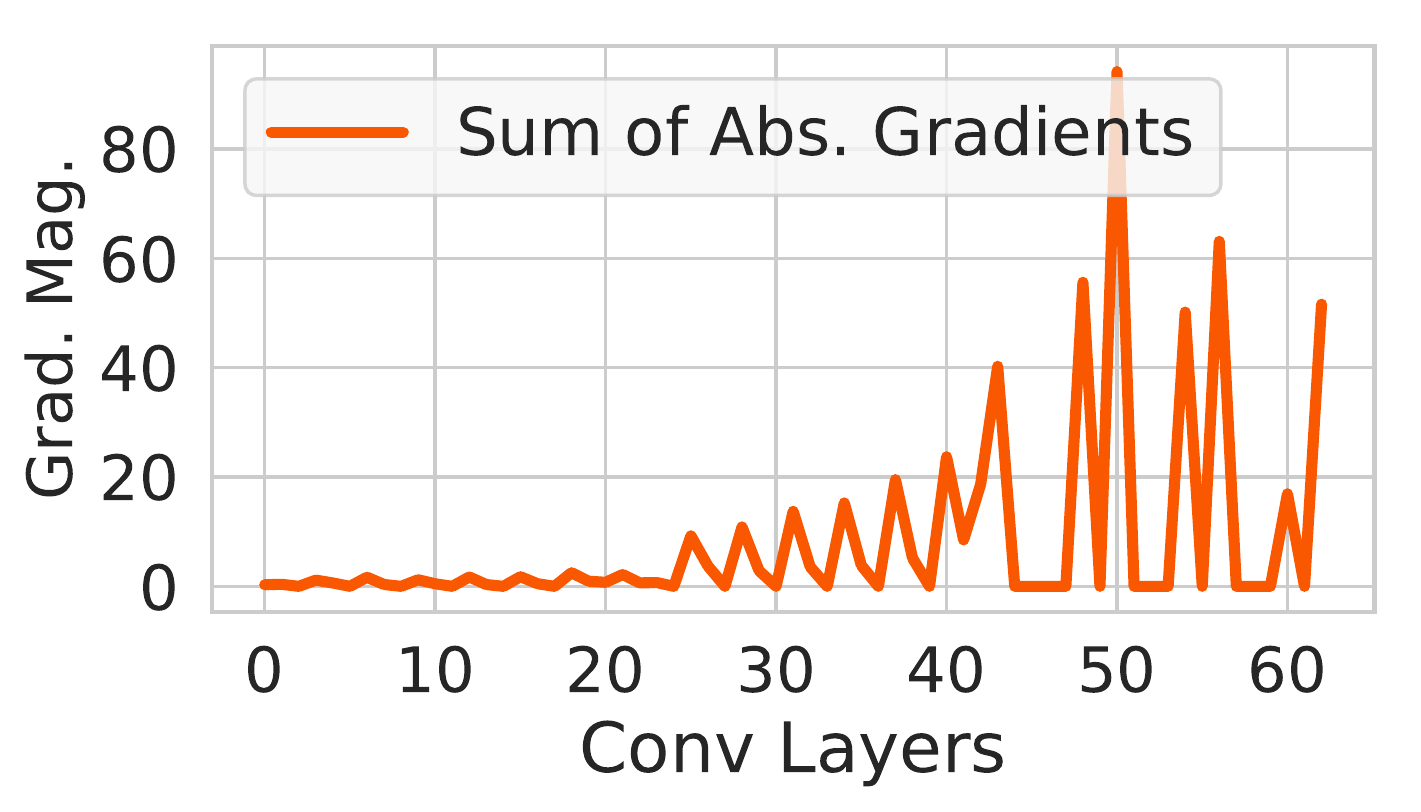}
    \includegraphics[width=0.28\textwidth]{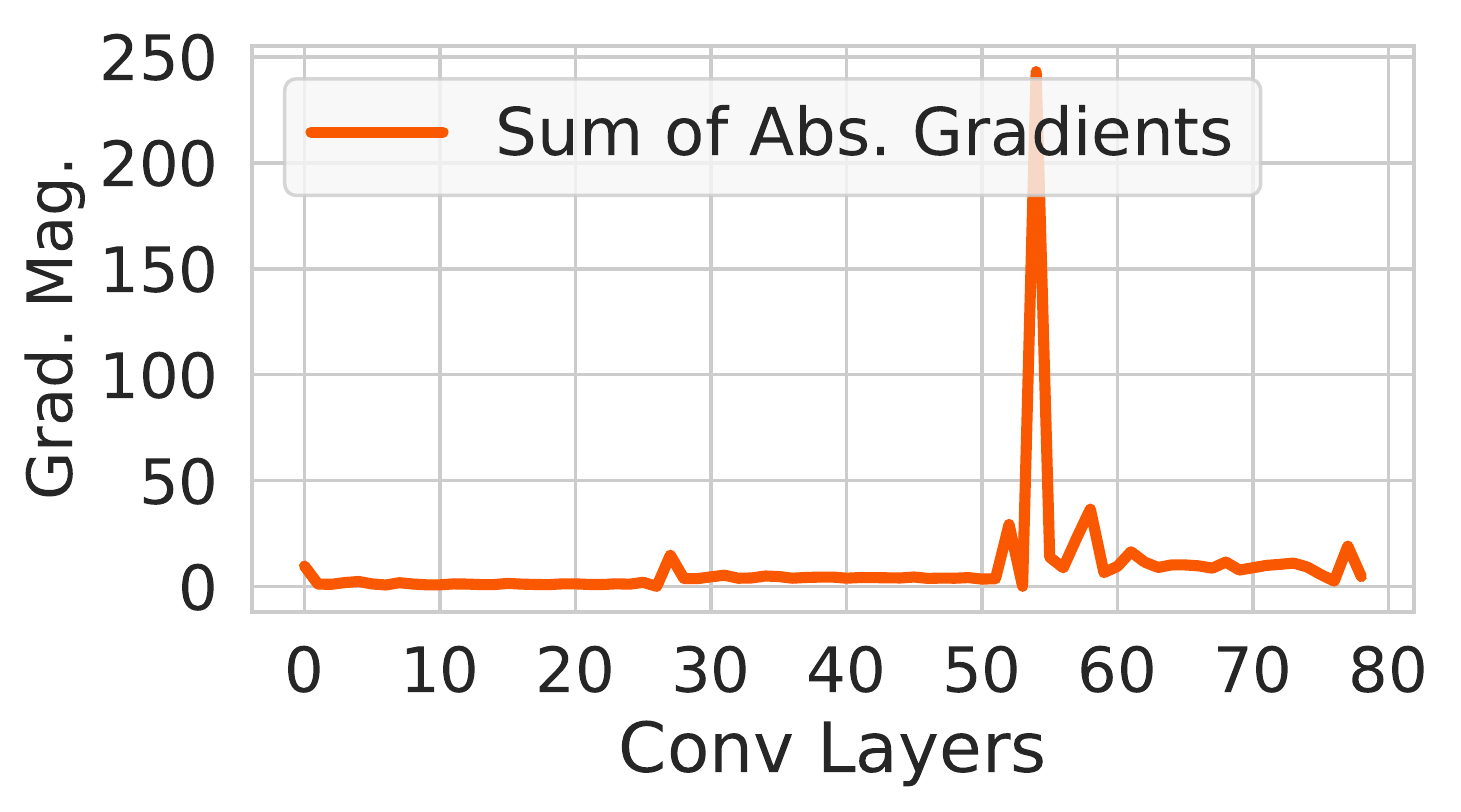}\\
    \vspace{-0.5em}
    {\small \hspace{2em}(a) Learning curve \hspace{5em}(b) Gradients w/o SC \hspace{1em}(c) Gradients w/ SC}
    \vspace{-1em}
    \caption{\textbf{Unstable gradients in the binary domain}. `w/o SC' indicates the cell template of~\cite{Zoph_2018_CVPR} (Fig.~\ref{fig:inter_cell}-(a)). `w/ SC' indicates the proposed cell template (Fig.~\ref{fig:inter_cell}-(b)). The gradient magnitudes are taken at epoch 100.
    With the proposed cell template (`w/ SC'), the searched network trains well (a).
    The proposed cell template shows far less spiky gradients along with generally larger gradient magnitudes ((b) vs. (c)), indicating that our template helps to propagate gradients more effectively in the binary domain}
    \label{fig:unstable_grads}
    \vspace{-1em}
\end{figure}

\begin{wrapfigure}{R}{0.5\textwidth}
\vspace{-3em}
\centering
\includegraphics[width=0.48\textwidth]{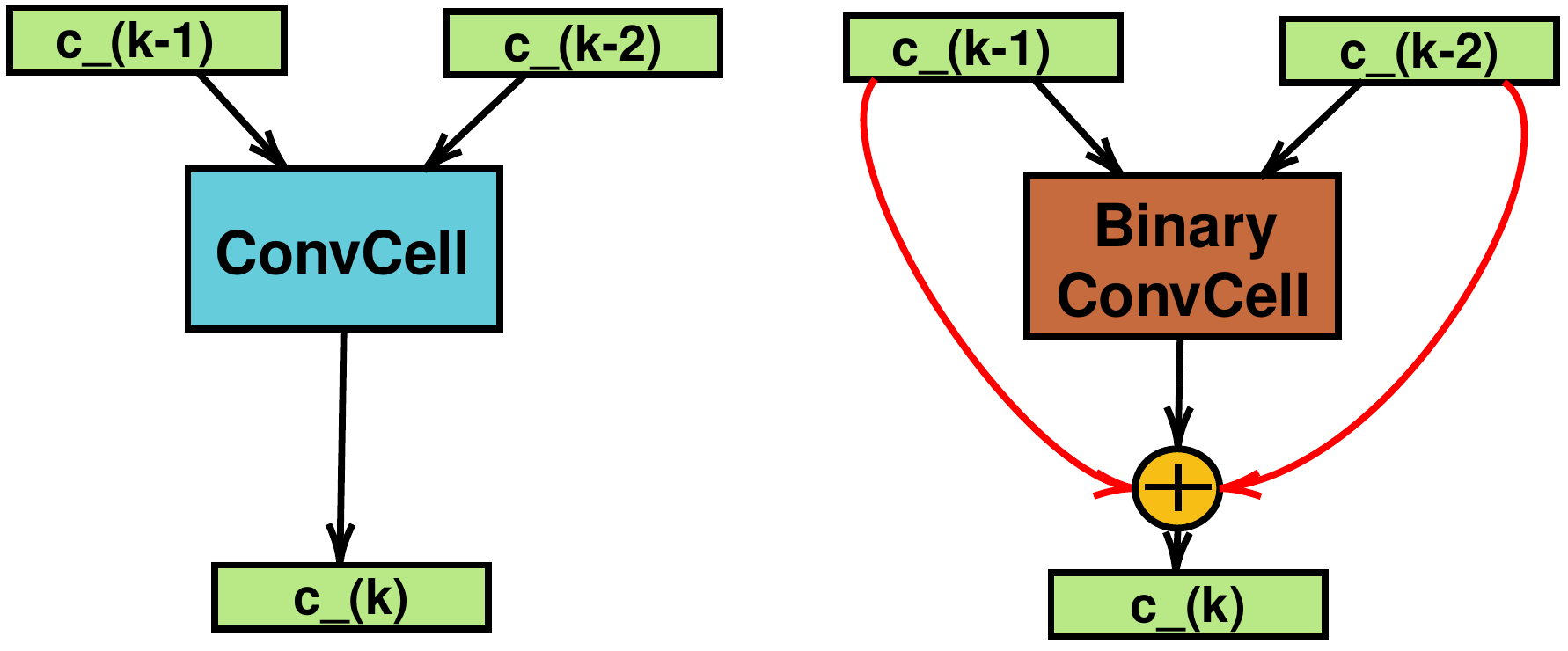}\\
\vspace{0.5em}
{\footnotesize (a) CT in DARTS~~~(b) CT in BNAS}
\vspace{-0.5em}
\caption{\textbf{Cell templates (CT) of (a) DARTS and (b) BNAS.} Red lines in BNAS indicate inter-cell skip connections. ConvCell indicates the convolutional cell. \textit{c\textunderscore(k)} indicates the output of the k\textsuperscript{th} cell}
\label{fig:inter_cell}
\vspace{-4em}
\end{wrapfigure}

With the defined search space, we now learn a network architecture with the convolutional cell template proposed in \cite{Zoph_2018_CVPR}.
However, the learned architecture still suffers from unstable gradients in the binary domain as shown in Fig.~\ref{fig:unstable_grads}-(a) and (b).
Investigating the reasons for the unstable gradients, we observe that the skip-connections in the cell template proposed in \cite{Zoph_2018_CVPR} are confined to be inside a single convolutional cell, \textit{i.e.}, intra-cell skip-connections.
The intra-cell skip-connections do not propagate the gradients outside the cell, forcing the cell to aggregate outputs that always have quantization error created inside the cell.
To help convey information without the cumulative quantization error through multiple cells, we propose to add skip-connections between multiple cells as illustrated in Fig.\ref{fig:inter_cell}.

\begin{table}[t!]
\centering
\caption{\textbf{Proposed search space for BNAS.} \textit{Bin Conv}, \textit{Bin Dil. Conv}, \textit{MaxPool} and \textit{AvgPool} refer to the binary convolution, binary dilated convolution, max pooling and average pooling layers, respectively}
\vspace{-1em}
\resizebox{0.85\linewidth}{!}{
\begin{tabular}{cccccccc}
\toprule
Layer Type                   & \multicolumn{2}{c}{Bin Conv.} & \multicolumn{2}{c}{Bin Dil. Conv.} & MaxPool  &AvgPool & Zeroise\\ 
\cmidrule(lr){1-1} \cmidrule(lr){2-3} \cmidrule(lr){4-5} \cmidrule(lr){6-6} \cmidrule(lr){7-7} \cmidrule(lr){8-8}
Kernel Size                  & $3\times3$         & $5\times5$         & $3\times3$            & $5\times5$           & $3\times3$            &$3\times3$ & N/A                      \\ 
\bottomrule
\end{tabular}
}
\vspace{-1em}
\label{table:layer_types}
\end{table}

The proposed cell template with inter-cell skip-connections help propagate gradients with less quantization error throughout the network, stabilizing the training curve.
Note that since the hand-crafted binary networks using the ResNet family as their backbone have similar residual connections (refer to the red line in Fig. \ref{fig:searched_cell_vs_xnor_resnet}).
We empirically validate the usefulness of the inter-cell skip connections in Sec.~\ref{sec:ablation}.

\vspace{-1.0em}
\subsection{Search Objective with Diversity Regularizer ($\widetilde{\mathcal{L}}_{S}$)}
\label{sec:diver_search}

With the feasible set of binary architectures ($A_B$) defined by $S_B$ and $T_B$, we solve the optimization problem similar to~\cite{liu2018darts}.
However, the layers with learnable parameters (\textit{e.g.}, convolutional layers) are not selected as often early on as the layers requiring no learning, because the parameter-free layers are more favorable than the under-trained layers.
The problem is more prominent in the binary domain because binary layers train slower than the floating point counterparts~\cite{courbariaux2016binarized}.
To alleviate this, we propose to use an exponentially annealed entropy based regularizer in the search objective to promote selecting diverse layers and call it the \textit{diversity regularizer}.
Specifically, we subtract the entropy of the architecture parameter distribution from the search objective as:
\begin{equation}
  \widetilde{\mathcal{L}}_{S}(D; \theta_{\alpha_B})  = \mathcal{L}_{S} (D;\theta, p) - \lambda H(p) e^{\left( -t/\tau \right)}, 
  \label{eq:diversity}
\end{equation}
where $\mathcal{L}_{S}(\cdot)$ is the search objective of \cite{liu2018darts}, which is a cross-entropy, $\theta_{\alpha_B}$ is the parameters of the sampled binary architecture, which is split into the architecture parameters $p$ and the network weights $\theta$, $H(\cdot)$ is the entropy, $\lambda$ is a balancing hyper-parameter, $t$ is the epoch, and $\tau$ is an annealing hyper-parameter.
This will encourage the architecture parameter distribution to be closer to uniform in the early stages, allowing the search to explore diverse layer types. 

Using the proposed diversity regularizer, we observed a $16\%$ relative increase in the average number of learnable layer types selected in the first $20$ epochs of the search.
More importantly, we empirically validate the benefit of the diversity regularizer with the test accuracy on the CIFAR10 dataset in Table~\ref{table:diversity_early} and in Sec.~\ref{sec:ablation}. 
While the accuracy improvement from the diversity regularizer in the floating point NAS methods such as DARTS \cite{liu2018darts} is marginal ($+0.2\%$), the improvement in our binary network is more meaningful ($+1.75\%$).

\begin{table}[t!]
\centering
\caption{\textbf{Effect of searching diversity on CIFAR10.} \textit{Diversity} refers to whether diversity regularization was applied ({\color{ForestGreen}\cmark}) or not ({\color{red}\xmark}) during the search. DARTS only gains $0.20\%$ test accuracy while BNAS gains $1.75\%$ test accuracy}
\vspace{-0.5em}
\begin{tabular}{ccccccc}
\toprule
Precision           & \multicolumn{3}{c}{Floating Point (DARTS)} & \multicolumn{3}{c}{Binary (BNAS)} \\
Diversity  & \color{red}\xmark    &  \color{ForestGreen}\cmark  & Gain   & \color{red}\xmark      & \color{ForestGreen}\cmark  & Gain \\ 
\cmidrule(lr){1-1} \cmidrule(lr){2-4} \cmidrule(lr){5-7}
 Test Acc. (\%)             &    $96.53$ & $96.73$ & {\color{ForestGreen}+$0.20$}      &  $90.95$&     $92.70$ & {\color{ForestGreen}+$1.75$} \\ 
\bottomrule
\end{tabular}
\vspace{-1em}
\label{table:diversity_early}
\end{table}

\section{Experiments}
\label{sec:exp}
\vspace{-0.5em}
\subsection{Experimental Setup}
\noindent\textbf{Datasets.}
We use CIFAR10\cite{krizhevsky09} and ImageNet (ILSVRC 2012)\cite{russakovsky2015imagenet} datasets to evaluate the image classification accuracy.
For searching binary networks, we use the CIFAR10 dataset. 
For training the final architectures from scratch, we use both CIFAR10 and ImageNet.
During the search, we hold out half of the training data of CIFAR10 as the validation set to evaluate the quality of search. 
For final evaluation of the searched architecture, we train it from the scratch using the full training set and report Top-1 (and Top-5 for ImageNet) accuracy.  

\vspace{0.5em}
\noindent\textbf{Details on Searching Architectures.}
We train a small network with $8$ cells and $16$ initial number of channels using SGD with the diversity regularizer (Sec.~\ref{sec:diver_search}) for $50$ epochs with batch size of $64$. 
We use momentum $0.9$ with initial learning rate of $0.025$ using cosine annealing \cite{loshchilov2016sgdr} and a weight decay of $3\times10^{-4}$. 
We use the same architecture hyper-parameters as \cite{liu2018darts} except for the additional diversity regularizer where we use $\lambda = 1.0$ and $\tau = 7.7$.
Our cell search takes approximately 10 hours on a single NVIDIA GeForce RTX 2080Ti GPU.

\vspace{0.5em}
\noindent\textbf{Details on Training the Searched Architectures.}
For CIFAR10,  we train the final networks for $600$ epochs with batch size $256$. 
We use SGD with momentum $0.9$ and weight decay of $3\times10^{-6}$.
We use the one cycle learning rate scheduler\cite{smith2018disciplined} with the learning rate ranging from 5 $\times$ 10\textsuperscript{-2} to 4 $\times$ 10\textsuperscript{-4}.
For ImageNet, we train the models for $250$ epochs with batch size $512$. 
We use SGD with momentum $0.9$, with an initial learning rate of $0.1$ and a weight decay  of $3\times10^{-5}$. 
We use the cosine restart scheduler \cite{loshchilov2016sgdr} with the minimum learning rate of $0$ and the length of one cycle being $50$ epochs.

\vspace{0.5em}
\noindent\textbf{Final Architecture Configurations.}

\begin{wraptable}{R}{0.5\textwidth}
\centering
\vspace{-3em}
\caption{\textbf{Configuration details of BNAS variants.} \emph{\# Cells}: the number of stacked cells. \emph{\# Chn.}: the number of output channels of the first convolution layer of the model. $\gamma$ is the transferrability hyper-parameter in Eq.~\ref{eq:zeroise}}
\resizebox{0.98\linewidth}{!}{
\label{table:networks}
\begin{tabular}{cccccccccc}
\toprule
BNAS- & Mini & A & B  & C & D & E & F & G & H\\ \midrule
\# Cells     & $10$       &  $20$  &   $12$ &  $16$ & $12$ & $12$ & $15$ &$11$ & $16$   \\
\# Chn.    & $24$ &   $36$       &    $64$      &    $108$ & $64$ & $68$ &$68$  &$74$ & $128$ \\
$\gamma$ &  $1$ & $1$       &      $1$    &    $1$ & $3$ & $3$ & $3$ & $3$ & $3$ \\ \midrule
Dataset    &\multicolumn{4}{c}{CIFAR10} & \multicolumn{5}{c}{ImageNet}   \\ 
\bottomrule
\end{tabular}
}
\vspace{-2em}
\end{wraptable}   

We vary the size of our BNAS to compare with the other binary networks with different FLOPs by stacking the searched cells and changing the output channels of the first convolutional layer and name them as BNAS-\{Mini, A, B, C, D, E, F, G, H\} as shown in Table~\ref{table:networks}.

\vspace{0.5em}
\noindent\textbf{Details on Comparison with Other Binary Networks.} 
For XNOR-Net with different backbone architectures, we use the floating point architectures from \verb torchvision ~or a public source\footnote{\href{https://github.com/kuangliu/pytorch-cifar}{https://github.com/kuangliu/pytorch-cifar}.} and apply the binarization scheme of XNOR-Net.
Following previous work \cite{liu2018bi, gu2019projection} on comparing ABC-Net with a single base~\cite{lin2017towards}, we compare PCNN with a single projection kernel for both CIFAR10 and ImageNet.

We will publicly release the code and learned models soon.

\subsection{Qualitative Analysis of the Searched Cell}
\label{sec:comp_search_hand_cell}

\begin{wrapfigure}{R}{0.5\textwidth}
\vspace{-3em}
\begin{center}
\includegraphics[width=0.5\columnwidth]{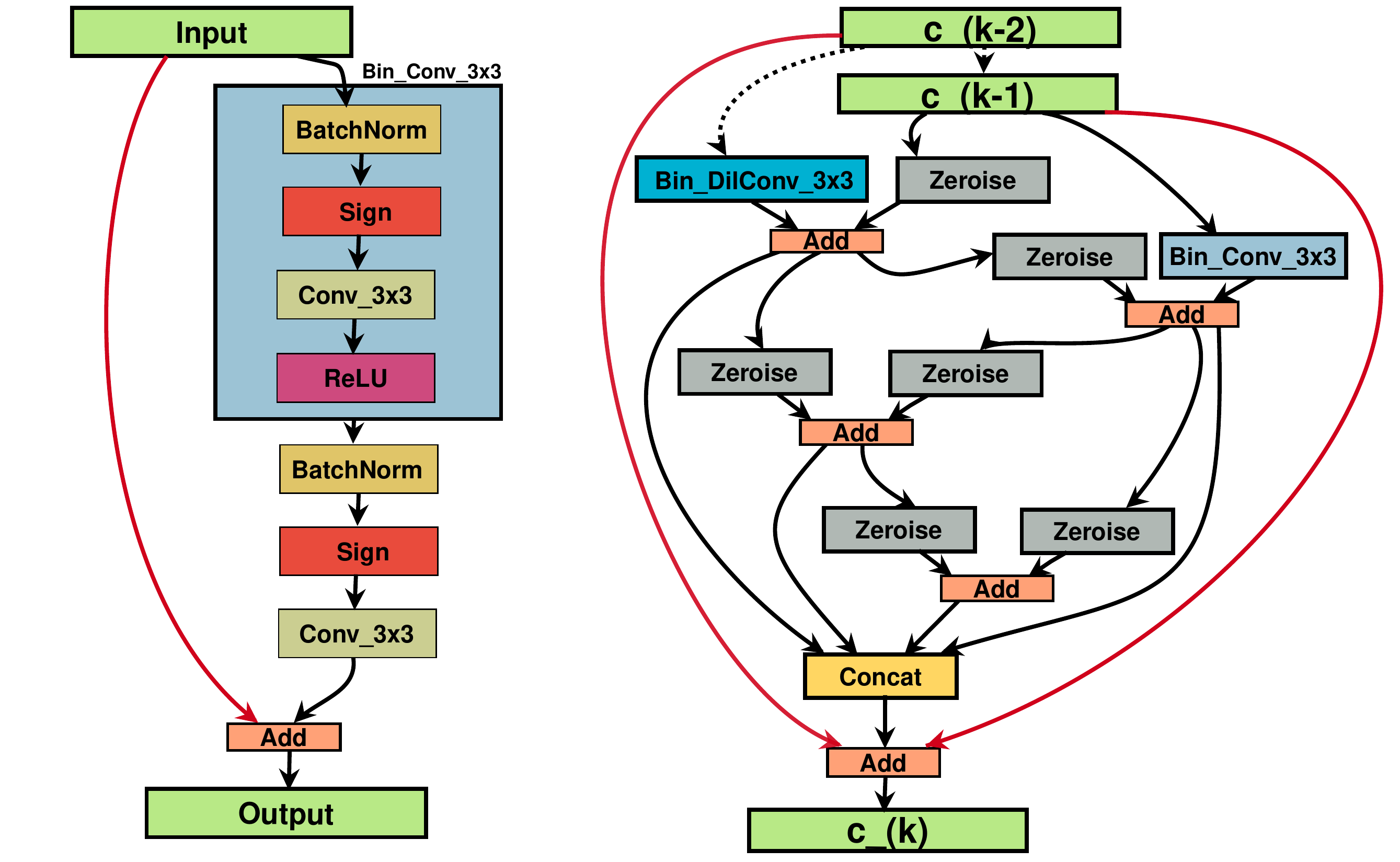}\\
\end{center}
\vspace{-1em}
\caption{\textbf{Comparing BNAS cell (right) to XNOR-Net cell (left).} \textit{c\textunderscore(k)} denotes the output of the k\textsuperscript{th} cell. The dotted lines represents the connections from the second previous cell (\textit{c\textunderscore(k-2)}). The red lines correspond to skip connections}
\vspace{-2em}
\label{fig:searched_cell_vs_xnor_resnet}
\end{wrapfigure}

We now qualitatively compare our searched cell with the XNOR-Net cell based on the ResNet18 architecture in Fig.~\ref{fig:searched_cell_vs_xnor_resnet}. 
As shown in the figure, our searched cell has a contrasting structure to the handcrafted ResNet18 architecture. 
Both cells contain only two 3$\times$3 binary convolution layer types, but the extra \textit{Zeroise} layer types selected by our search algorithm help in reducing the quantization error. 
The topology in which the \textit{Zeroise} layer types and convolution layer types are connected also contributes to improving the classification performance of our searched cell.
In the following subsections, we show that our searched topology yields better binary networks that outperform the architectures used in state-of-the-art binary networks.
More qualitative comparisons of our searched cell can be found in the supplement.

\begin{table}[t!]
\centering
\caption{\textbf{Comparison of different backbone architectures for binary networks with XNOR-Net binarization scheme \cite{Rastegari2016XNORNetIC} in various FLOPs budgets.} 
Bi-Real* indicates Bi-Real Net's method with only the architectural modifications. 
We refer to \cite{Kim2020BinaryDuo} for the FLOPs of CBCN. 
CBCN* indicates the highest accuracy for CBCN with the ResNet18 backbone as \cite{Liu2019CirculantBC} report multiple different accuracy for the same network configuration. Additionally, \cite{Liu2019CirculantBC} does not report the exact FLOPs of their model, hence we categorized them conservatively into the `$\sim0.27$' bracket (we also outperform them with a smaller model: BNAS-A)}
\vspace{-0.5em}
\resizebox{0.96\linewidth}{!}{
\begin{tabular}{ccccc}
\toprule
Dataset & FLOPs ($\times10^8$)                & Model (Backbone Arch.)             & Top-1 Acc. (\%) & Top-5 Acc. (\%)\\ \midrule
\multirowcell{11}{\rotatebox[origin=c]{90}{CIFAR10}} & \multirow{5}{*}{$\sim 0.16$} & XNOR-Net (ResNet18) &  $88.82$ & -        \\
                      &      & XNOR-Net (DenseNet) &  $85.16$  & -\\
                      &      & XNOR-Net (NiN) &  $86.28$  & -\\
                      &      & XNOR-Net (SENet) &  $88.12$ & - \\ 
        &    & \cellcolor{Gray} BNAS-A             &  \cellcolor{Gray} ${\bf 92.70}$ & \cellcolor{Gray} -         \\ 
                            \cmidrule{2-5}
       & \multirow{4}{*}{$\sim 0.27$} 
                        & XNOR-Net (ResNet34)  &  $88.54$ & -\\ 
                        &     & XNOR-Net (WRN40)    &   $91.58$ & -\\ 
                        &     & CBCN* (ResNet18)~\cite{Liu2019CirculantBC} & $91.91$ & - \\
                        &      & \cellcolor{Gray} BNAS-B & \cellcolor{Gray} ${\bf 93.76}$ & \cellcolor{Gray} -         \\ 
                            \cmidrule{2-5}
&\multirow{2}{*}{$\sim 0.90$} & XNOR-Net (ResNext29-64) &  $84.27$ & -\\ 
 &                              &\cellcolor{Gray}  BNAS-C  & \cellcolor{Gray} ${\bf 94.43}$ & \cellcolor{Gray} -\\
                            \midrule
\multirowcell{11}{\rotatebox[origin=c]{90}{ImageNet}}
 & \multirow{2}{*}{$\sim 1.48$} & XNOR-Net (ResNet18) & $51.20$ & $73.20$ \\ 
 &                              &\cellcolor{Gray}  BNAS-D &\cellcolor{Gray}  ${\bf 57.69}$ &\cellcolor{Gray}  ${\bf 79.89}$ \\
                                \cmidrule{2-5}
 & \multirow{2}{*}{$\sim 1.63$} & Bi-Real* (Bi-Real Net18)~\cite{liu2018bi}&  $32.90$ & $56.70$ \\ 
&                               &\cellcolor{Gray} BNAS-E& \cellcolor{Gray} $\bf 58.76$  &\cellcolor{Gray} $\bf 80.61$  \\ 
                               \cmidrule{2-5}
&\multirow{2}{*}{$\sim 1.78$} & XNOR-Net (ResNet34) & $56.49$  & $79.13$\\ 
 &                            &\cellcolor{Gray}  BNAS-F & \cellcolor{Gray}  ${\bf 58.99}$  & \cellcolor{Gray}  ${\bf 80.85}$\\ 
                              \cmidrule{2-5}  
&\multirow{2}{*}{$\sim 1.93$} & Bi-Real* (Bi-Real Net34)~\cite{liu2018bi} & $53.10$ & $76.90$ \\ 
        &   &\cellcolor{Gray} BNAS-G &\cellcolor{Gray} ${\bf 59.81}$&\cellcolor{Gray} ${\bf 81.61}$\\
                                    \cmidrule{2-5}
&\multirow{2}{*}{$\sim 6.56$}& CBCN (Bi-Real Net18)~\cite{Liu2019CirculantBC} & $61.40$ & $82.80$ \\ 
        &   &\cellcolor{Gray} BNAS-H & \cellcolor{Gray} $\bf 63.51$ &\cellcolor{Gray} $\bf 83.91$ \\
\bottomrule
\end{tabular}
}
\vspace{-1em}
\label{table:comp_arch}
\end{table}

\vspace{-1em}
\subsection{Comparisons on Backbone Architectures for Binary Networks}
\label{sec:comp_arch}

We now quantitatively compare our searched architectures to various backbone architectures that have been used in the state-of-the-art binary networks with the binarization scheme of XNOR-Net~\cite{Rastegari2016XNORNetIC} in Table~\ref{table:comp_arch}.
The comparisons differ only in the backbone architecture, allowing us to isolate the effect of our searched architectures on the final accuracy, \textit{i.e} the comparison with XNOR-Net with different backbone architectures for various FLOPs and newer binary networks with the architectural contributions only. 
To single out the architectural contributions of Bi-Real Net, we used Table 1 in~\cite{liu2018bi} to excerpt the ImageNet classification accuracy with using only the Bi-Real Net architecture. Note that CBCN is based on the Bi-Real Net architecture with the convolutions being changed to circulant convolutions\footnote{They mention that center loss and gaussian gradient update is also used but they are not elaborated and not the main focus of CBCN's method.}.
Additionally, as mentioned in Sec.~\ref{sec:related}, we do not compare with \cite{shen2019searching} as the inference speed-up is significantly worse than other binary networks ($\sim 2.7\times $ compared to $\sim10\times$), which makes the comparison less meaningful.

As shown in Table~\ref{table:comp_arch}, our searched architectures outperform other architectures used in binary networks in all FLOPs brackets and on both CIFAR10 and ImageNet.
Notably, comparing XNOR-Net with the ResNet18 and ResNet34 backbone to BNAS-D and BNAS-F, we gain $+6.49\%$ or $+2.50\%$ top-1 accuracy and $+6.69\%$ or $+1.72\%$ top-5 accuracy on ImageNet.

Furthermore, BNAS retains the accuracy much better at lower FLOPs, showing that our searched architectures are better suited for efficient binary networks.
Additionally, comparing CBCN to BNAS-H, we gain $+2.11\%$ top-1 accuracy and $+1.11\%$ top-5 accuracy, showing that our architecture can scale to higher FLOPs budgets better than CBCN.
Although binary networks are more useful in lesser FLOPs budgets, if one want to further improve the scalability of our searched architectures to higher FLOPs budgets, a grid search for the hyper-paramters shown in Table~\ref{table:networks} can be performed similar to~\cite{Tan2019EfficientNetRM}.
In sum, replacing the architectures used in current binary networks to our searched architectures can greatly improve the performance of binary networks.

\vspace{-1em}
\subsection{Comparison with Other Binary Networks}
\label{sec:comp_sota}

\begin{table}[t!]
\centering
\caption{\textbf{Comparison of other binary networks in various FLOPs budgets.} The binarization schemes are: `\textit{Sign + Scale}': using fixed scaling factor and the sign function~\cite{Rastegari2016XNORNetIC}, `\textit{Sign}': using the sign function~\cite{courbariaux2016binarized}, `\textit{Clip + Scale}': using clip function with shift parameter~\cite{lin2017towards}, , `\textit{Sign + Scale*}': using learned scaling factor and the sign function~\cite{bulat2019xnor}, `\textit{Projection}': using projection convolutions\cite{gu2019projection}, `\textit{Bayesian}': using a learned scaling factor from the Bayesian losses~\cite{gu2019bayesian} and the sign function, and `\textit{Decoupled}': decoupling ternary activations to binary activations~\cite{Kim2020BinaryDuo}}
\vspace{-0.5em}
\resizebox{0.98\linewidth}{!}{
\begin{tabular}{ccccccc}
\toprule
Dataset & FLOPs ($\times10^8$)                & Method (Backbone Arch.)     & Binarization Scheme &   Pretraining & Top-1 Acc. (\%) & Top-5 Acc. (\%)\\ \midrule
\multirowcell{7}{\rotatebox[origin=c]{90}{CIFAR10}} 
                & \multirow{2}{*}{$\sim 0.04$} 
                    & PCNN($i=16$) (ResNet18)~\cite{gu2019projection} & Projection & \color{red}\xmark &$89.16$ & -        \\ 
                &               & \cellcolor{Gray} BNAS-Mini &\cellcolor{Gray}  Sign + Scale &\cellcolor{Gray}  \color{red}\xmark &\cellcolor{Gray}  ${\bf 90.12}$ &\cellcolor{Gray}  - \\ 
                \cmidrule{2-7}
&\multirow{2}{*}{$\sim 0.16$} 
                            & BinaryNet (ResNet18)~\cite{courbariaux2016binarized} & Sign& \color{red}\xmark & $89.95$ & -\\  
     &      & \cellcolor{Gray} BNAS-A & \cellcolor{Gray} Sign + Scale & \cellcolor{Gray} \color{red}\xmark  & \cellcolor{Gray} ${\bf 92.70}$ & \cellcolor{Gray} -         \\ 
                            \cmidrule{2-7}
&\multirow{2}{*}{$\sim 0.27$} & PCNN($i=64$) (ResNet18)~\cite{gu2019projection} & Projection & \color{ForestGreen}\cmark & ${\bf 94.31}$ & -\\ 
  &         &\cellcolor{Gray} BNAS-B &\cellcolor{Gray} Sign + Scale &\cellcolor{Gray} \color{red}\xmark  &\cellcolor{Gray} $93.76$ &\cellcolor{Gray} -\\
                            \midrule
\multirowcell{14}{\rotatebox[origin=c]{90}{ImageNet}} &\multirow{3}{*}{$\sim 1.48$} & BinaryNet (ResNet18)~\cite{courbariaux2016binarized} & Sign & \color{red}\xmark & $42.20$ & $67.10$ \\
 &                            & ABC-Net (ResNet18)~\cite{lin2017towards} & Clip + Sign& \color{red}\xmark & $42.70$ & $67.60$ \\ 
 &          &\cellcolor{Gray} BNAS-D &\cellcolor{Gray}Sign + Scale &\cellcolor{Gray} \color{red}\xmark &\cellcolor{Gray} ${\bf 57.69}$ &\cellcolor{Gray} ${\bf 79.89}$ \\
                             \cmidrule{2-7}
& \multirow{6}{*}{$\sim 1.63$} & Bi-Real (Bi-Real Net18)~\cite{liu2018bi} & Sign + Scale & \color{ForestGreen}\cmark &  $56.40$ & $79.50$ \\ 
                 &            & XNOR-Net++ (ResNet18)~\cite{bulat2019xnor} & Sign + Scale* & \color{red}\xmark & $57.10$ & $79.90$ \\ 
                 &            & PCNN (ResNet18)~\cite{gu2019projection} & Projection & \color{ForestGreen}\cmark & $57.30$ & $80.00$ \\ 
                 &            & BONN (Bi-Real Net18)~\cite{gu2019bayesian} & Bayesian & \color{red}\xmark & $59.30$ & $81.60$ \\ 
                 &            & BinaryDuo (ResNet18)~\cite{Kim2020BinaryDuo} & Decoupled & \color{ForestGreen}\cmark & ${\bf 60.40}$ & ${\bf 82.30}$ \\ 
     &      &\cellcolor{Gray}  BNAS-E &\cellcolor{Gray} Sign + Scale &\cellcolor{Gray} \color{red}\xmark &\cellcolor{Gray}  $58.76$  &\cellcolor{Gray} $80.61$  \\ 
                              \cmidrule{2-7}
&\multirow{2}{*}{$\sim 1.78$}  & ABC-Net (ResNet34)~\cite{lin2017towards} & Clip + Scale & \color{red}\xmark & $52.40$ & $76.50$ \\ 
   &        &\cellcolor{Gray} BNAS-F &\cellcolor{Gray} Sign+Scale&\cellcolor{Gray} \color{red}\xmark &\cellcolor{Gray} ${\bf 58.99}$ &\cellcolor{Gray} ${\bf 80.85}$\\
                              \cmidrule{2-7}
&\multirow{2}{*}{$\sim 1.93$}& Bi-Real (Bi-Real Net34)~\cite{liu2018bi} & Sign + Scale & \color{ForestGreen}\cmark &  ${\bf 62.20}$ & ${\bf 83.90}$ \\ 
        &   &\cellcolor{Gray} BNAS-G &\cellcolor{Gray} Sign + Scale &\cellcolor{Gray} \color{red}\xmark &\cellcolor{Gray} $59.81$&\cellcolor{Gray} $81.61$\\
\bottomrule
\end{tabular}
}
\vspace{-1em}
\label{table:comp_sota}
\end{table}

As we focus on improving binary networks by architectural benefits only, comparison to other binary network methods is not of our interest.
However, it is still intriguing to compare gains from a pure architectural upgrade to gains from new binarization schemes or new training methods.
As shown in Table~\ref{table:comp_sota}, our searched architectures outperform other methods in more than half the FLOPs brackets spread across CIFAR10 and ImageNet.
Moreover, the state-of-the-art methods that focus on discovering better training schemes are complementary to our searched architectures, as these training methods were not designed exclusively for a fixed network topology.

Note that, with the same backbone of ResNet18 or ResNet34, Bi-Real, PCNN, XNOR-Net++ and BONN have higher FLOPs than ABC-Net, XNOR-Net and BinaryNet.
The higher FLOPs are from unbinarizing the downsampling convolutions in the ResNet architecture.\footnote{We have confirmed with the authors of \cite{Rastegari2016XNORNetIC} that their results were reported without unbinarizing the downsampling convolutions.}

\subsection{Ablation Studies}
\label{sec:ablation}

We perform ablation studies on the proposed components of our method.
We use the CIFAR10 dataset for the experiments with various FLOPs budgets and summarize the results in Table~\ref{table:ablation}.

\begin{wraptable}{R}{0.59\textwidth}
\vspace{-3em}
\centering
\caption{\textbf{Classification accuracy (\%) of ablated models on CIFAR10.} \textit{Full} refers to the proposed method with all components. \textit{No Skip} refers to our method without the inter-cell skip connections. \textit{No Zeroise} refers to our method with explicitly discarding the \textit{Zeroise} layers. \textit{No Div} refers to our method without the diversity regularizer.}
\vspace{0.5em}
\resizebox{0.98\linewidth}{!}{
\begin{tabular}{ccccc}
\toprule
Model & Full & No Skip  & No Zeroise & No Div\\ \midrule
BNAS-A     & \cellcolor{Gray} ${\bf 92.70}$    & $61.23$    &   $89.47$   & $90.95$    \\ 
BNAS-B      &\cellcolor{Gray} ${\bf 93.76}$   & $67.15$      & $91.69$      & $91.55$  \\ 
BNAS-C    &\cellcolor{Gray} ${\bf 94.43}$ & $70.58$    &   $88.74$         & $92.66$ \\
\bottomrule
\end{tabular}
}
\vspace{-2em}
\label{table:ablation}
\end{wraptable}

All components have decent contributions to the accuracy, with the inter-cell skip connection in the new cell template contributing the most; without it, the models eventually collapsed to very low training and test accuracy and exhibited unstable gradient issues as discussed in Sec.~\ref{sec:cell_design}. Comparing \textit{No Div} with \textit{Full}, the searched cell with the diversity regularizer has a clear gain over the searched cell without it in all the model variants. 
Interestingly, the largest model (BNAS-C) without \textit{Zeroise} layers performs worse than BNAS-A and BNAS-B, due to excess complexity.
Please refer to the supplement for more discussion regarding the ablations.

\section{Conclusion}
\label{sec:conclusion}
To design better performing binary network architectures, we propose a method to search the space of binary networks, called BNAS.
BNAS searches for a cell that can be stacked to generate networks for various computational budgets.
To configure the feasible set of binary architectures, we define a new search space of binary layer types and a new cell template.
Specifically, we propose to exclude separable convolution layer and include \textit{Zeroise} layer type in the search space for less quantization error.
Further, we propose a new search objective with the diversity regularizer and show that it helps in obtaining better binary architectures.
The learned architectures outperform the architectures used in the state-of-the-art binary networks in the same computational budget in FLOPs on both CIFAR-10 and ImageNet.

\section*{Acknowledgement}

\noindent 
The authors would like to thank Dr. Mohammad Rastegari at XNOR.AI for valuable comments and training details of XNOR-Net and Dr. Chunlei Liu and other authors of \cite{Liu2019CirculantBC} for sharing the code of \cite{Liu2019CirculantBC}.

%
%
\bibliographystyle{splncs04}
\bibliography{bibs/bib}

\newpage
\appendix 
\section*{Appendix}

\section{Additional Qualitative Analysis of Our Searched Cell}
\label{sec:qual}

We present more qualitative comparisons of both the normal cell and the reduction cell of our searched cell with the binarized DARTS cell~\cite{liu2018darts}, in addition to Section~\ref{sec:comp_search_hand_cell} where we compare the normal cell with the hand-crafted XNOR-Net cell.

\begin{figure}[b!]
\centering
\includegraphics[width=0.3\columnwidth]{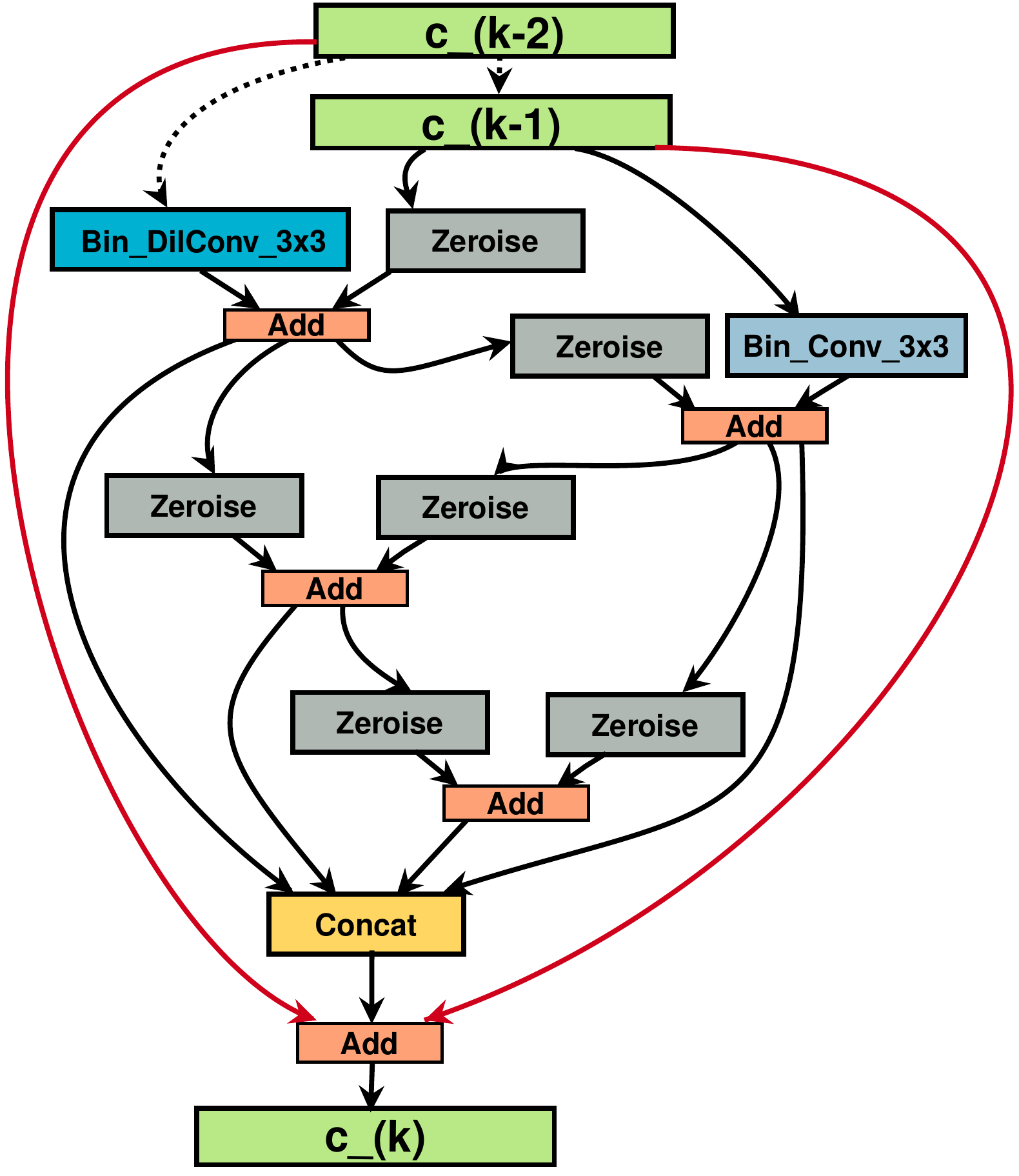}
\hspace{4em}
\includegraphics[width=0.43\columnwidth]{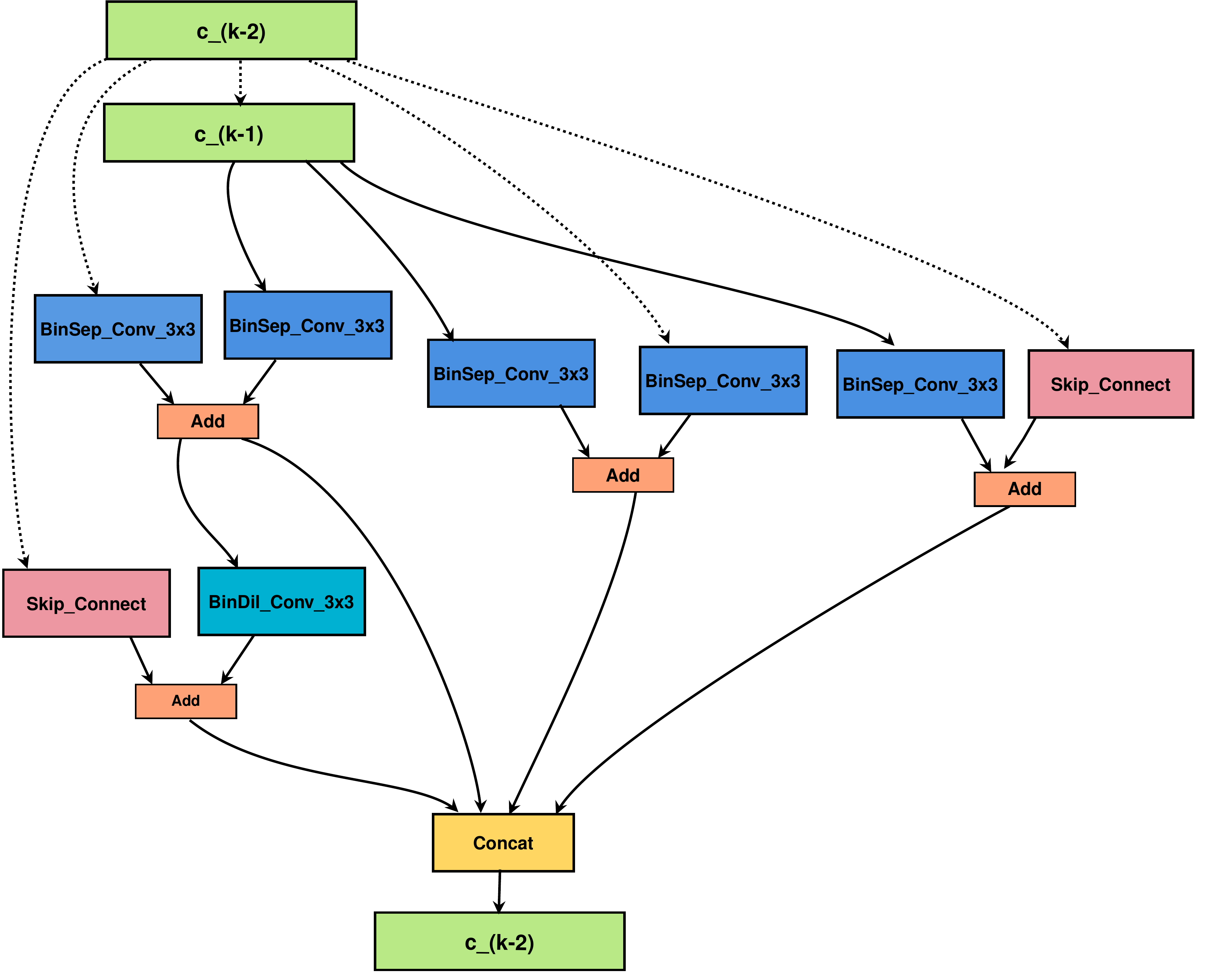}\\
\vspace{-.1em}
{\small (a) BNAS Normal Cell \hspace{5em}(b) Binarized DARTS Normal Cell} 
\vspace{0.5em}
\caption{\textbf{Comparing the normal cell of BNAS (a) and the normal cell of binarized DARTS (b).} \textit{c\textunderscore(k)} indicates the output of the $k^\text{th}$ cell. The dotted lines represent
the connections from the second previous cell (\textit{c\textunderscore(k-2)}). {\color{red}Red lines} in (a) indicate the inter-cell skip connections. Note that the searched cell of binarized DARTS in (b) has only intra-cell skip connections (denoted by \colorbox{pink!30}{pink boxes}) which have unstable gradients as compared to inter-cell skip connections in (a) (See discussions in Section~\ref{sec:cell_design})}
\vspace{-.5em}
\label{fig:compare_darts}
\end{figure}

\vspace{0.5em}
\noindent\textbf{Normal Cell.}
In Figure~\ref{fig:compare_darts}, we compare the normal cell of BNAS with the normal cell of DARTS\cite{liu2018darts}. 
Our cell has inter-cell skip connections which result in more stable gradients leading to better training, whereas the binarized DARTS cell does not train at all (achieving only 10.01\% test accuracy on CIFAR10 in Figure~\ref{fig:search_constraint} and Table~\ref{table:order_bin}). 
We hypothesize that the lack of inter-cell skip connections in their cell template may also contribute to the failure of its architecture in the binary domain other than the excessive number of separable convolutions in the DARTS searched cell (Table~\ref{table:order_bin}).

\begin{figure}[h!]
\centering
\includegraphics[width=0.345\columnwidth]{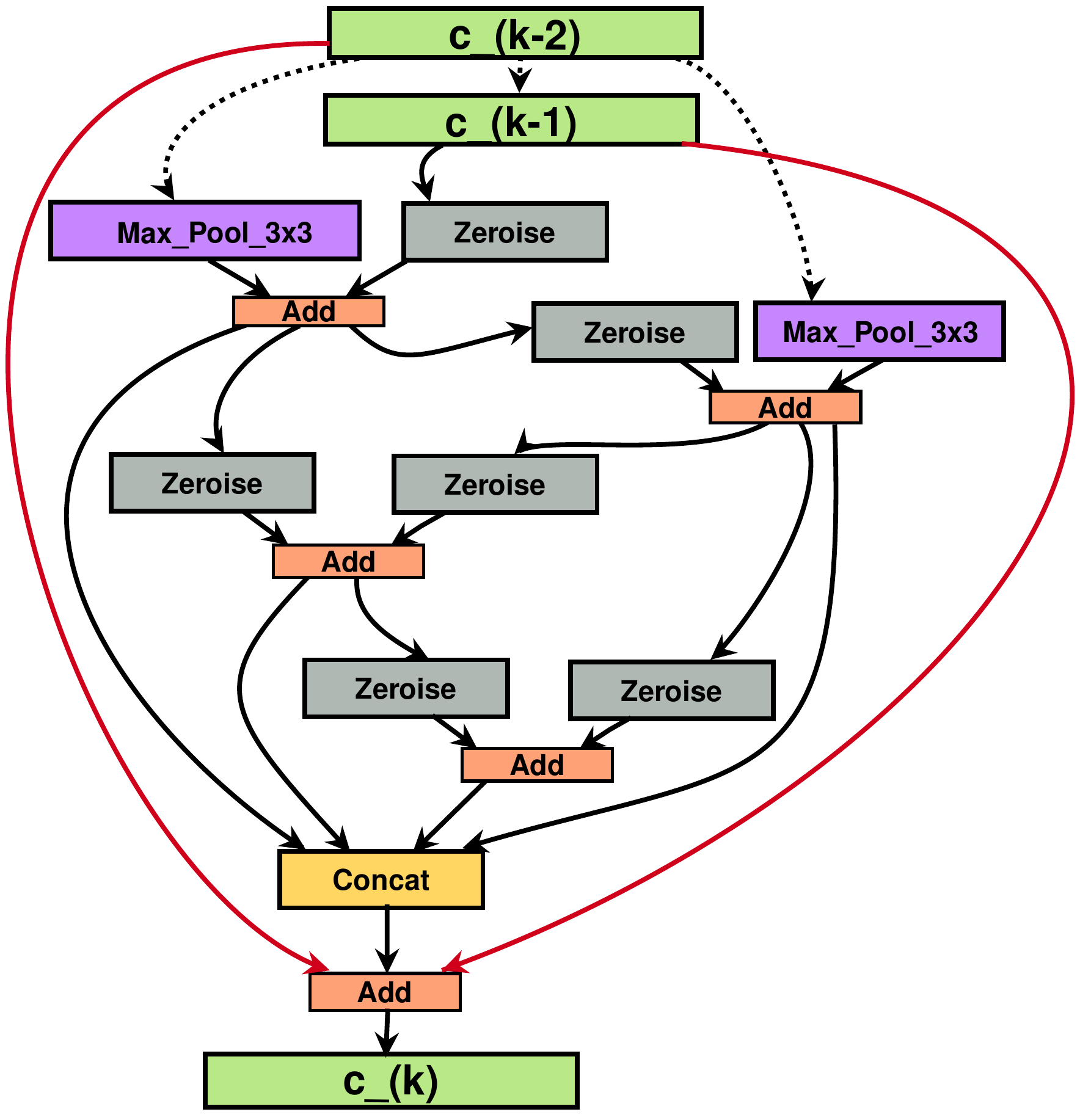}
\hspace{2em}
\includegraphics[width=0.45\columnwidth]{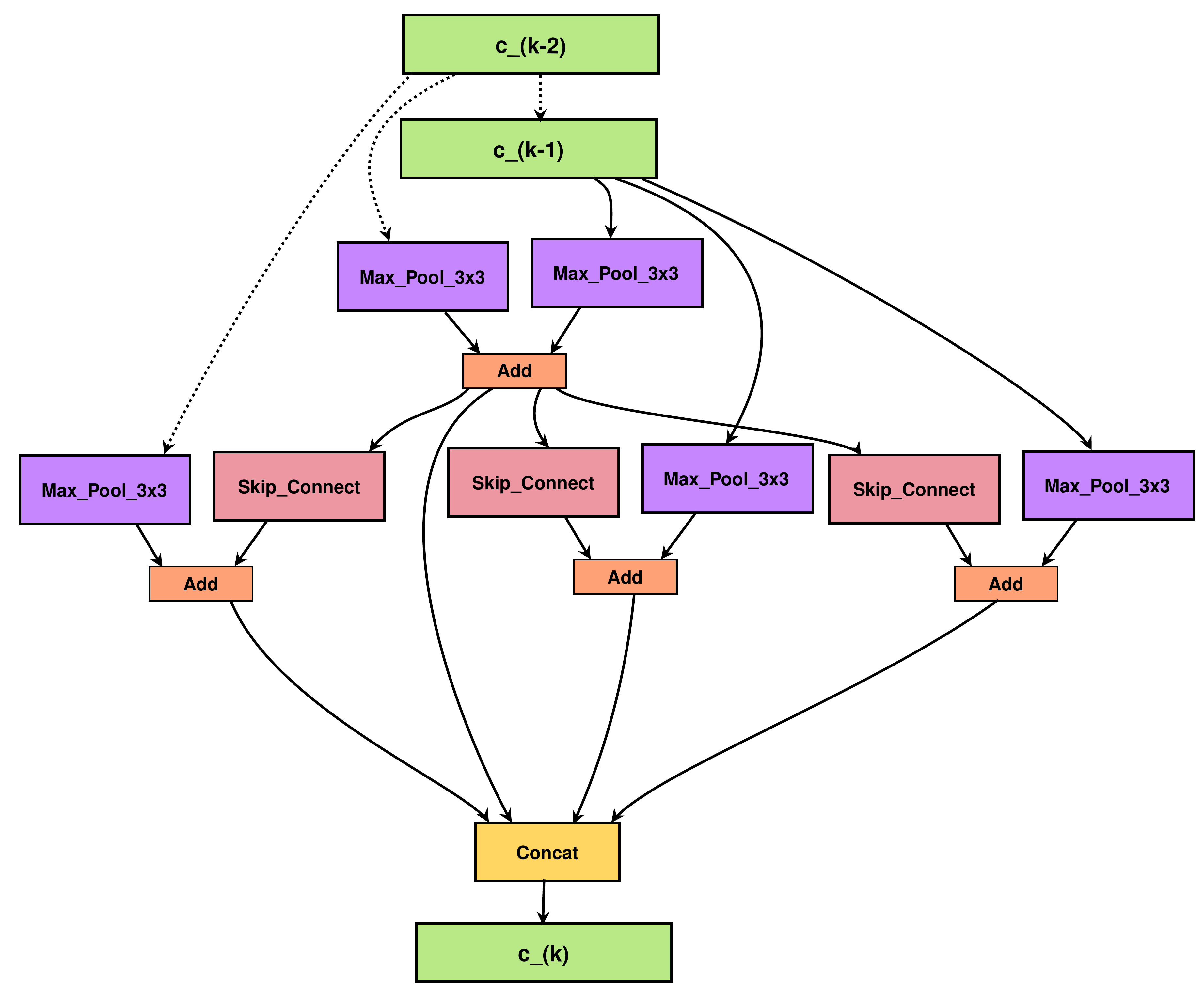}\\
\vspace{-.1em}
{\small(a) BNAS Reduction Cell \hspace{5em}(b) Binarized DARTS Reduction Cell} 
\vspace{0.5em}
\caption{\textbf{Comparing the reduction cell of BNAS (a) and the reduction cell of binarized DARTS (b)}. \textit{c\textunderscore(k)} indicates the output of the $k^\text{th}$ cell. The dotted lines represent the connections from the second previous cell (\textit{c\textunderscore(k-2)}). {\color{red}Red lines} indicate the inter-cell skip connections. The intra-cell skip connections are denoted by the \colorbox{pink!30}{pink boxes}. Interestingly, the BNAS reduction cell only uses the output from the second previous cell (\textit{c\textunderscore(k-2)}) as inputs to the max pool layers, utilizing the inter-cell skip connections more}
\vspace{-0.5em}
\label{fig:red_cell}
\end{figure}

\vspace{0.5em}
\noindent\textbf{Reduction Cell.}
We also qualitatively compare the BNAS reduction cell to the binarized DARTS reduction cell in Figure~\ref{fig:red_cell}. 
Note that the BNAS reduction cell has a lot of \textit{Zeroise} layers which help reduce quantization error.

\section{Additional Analyses on the Ablated Models}
\label{sec:detailed_ablation}

\subsection{`No Skip' Setting}
\label{sec:no_skip}

\begin{figure}[h!]
\centering
\includegraphics[width=0.4\columnwidth]{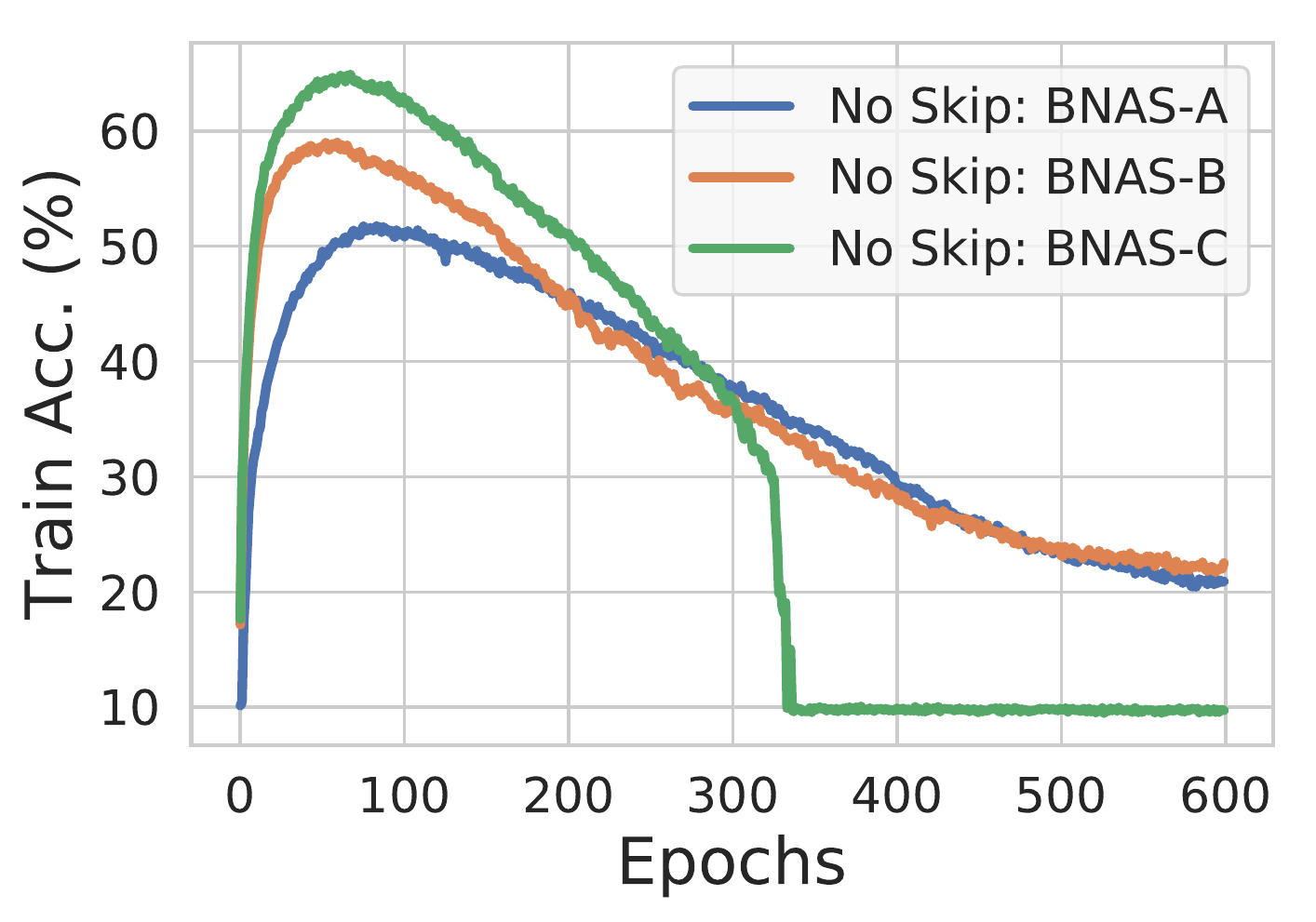}\hspace{3em}
\includegraphics[width=0.4\columnwidth]{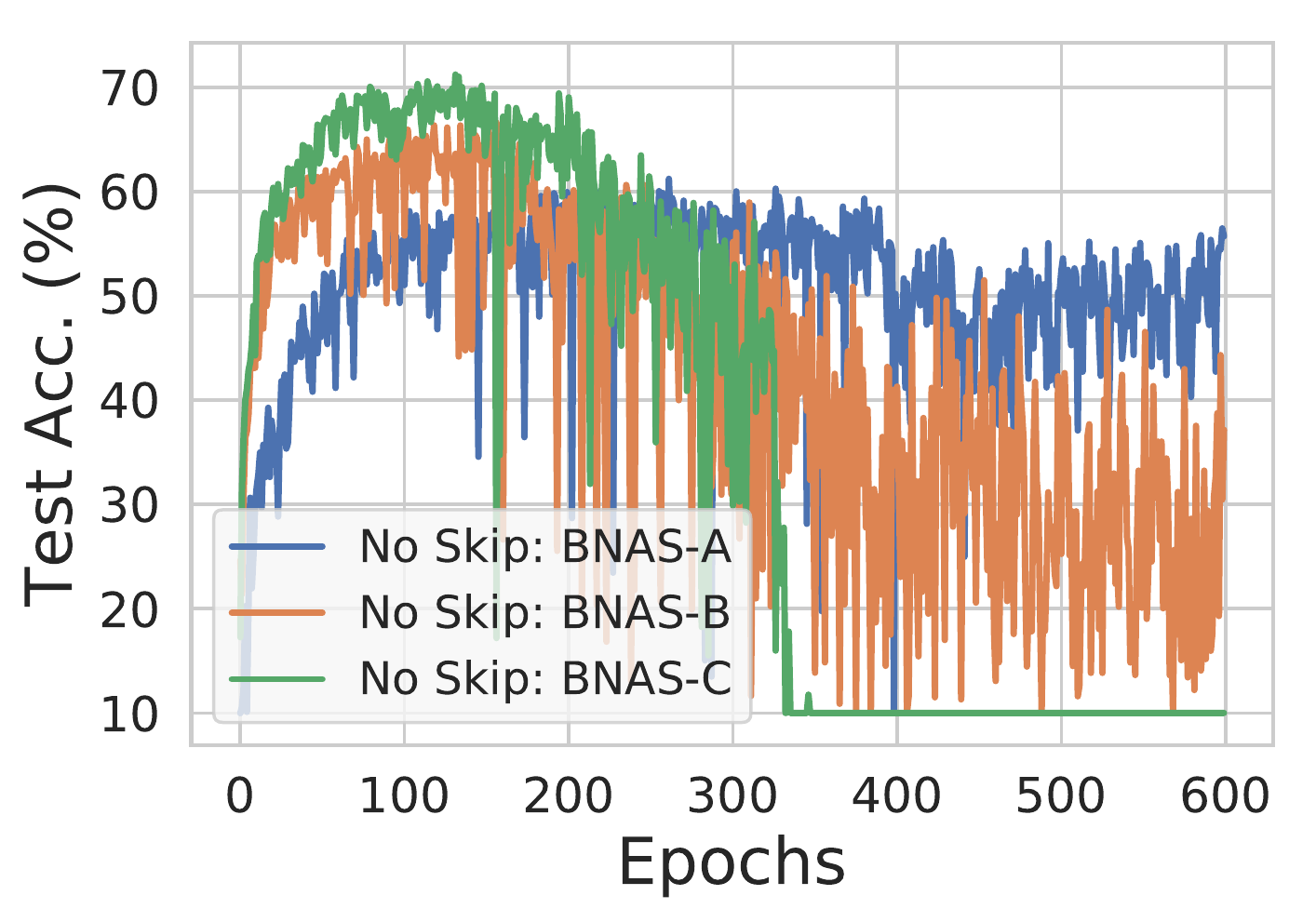}\\
\vspace{-.1em}
{\small (a) Train Accuracy (\%) on CIFAR10  \hspace{1.5em}(b) Test Accuracy (\%) on CIFAR10} 
\caption{\textbf{Learning curve for the `No Skip' ablation}. The train (a) and test (b) accuracy of all three models collapse when trained for 600 epochs. Additionally, the test accuracy curves fluctuates heavily when compared to the train accuracy curve}
\vspace{-0.5em}
\label{fig:no_skip_train_curve}
\end{figure}

Besides the final classification accuracy presented in Table~\ref{table:ablation}, here we additionally present the train and test accuracy curves for the \textit{No Skip} ablation models of Table~\ref{table:ablation} in Figure~\ref{fig:no_skip_train_curve} for more detailed analysis. 
All three variants collapse to a very low training and test accuracy after a reasonable number of epochs (600). 

\begin{figure}[b!]
\centering
\includegraphics[width=0.31\columnwidth]{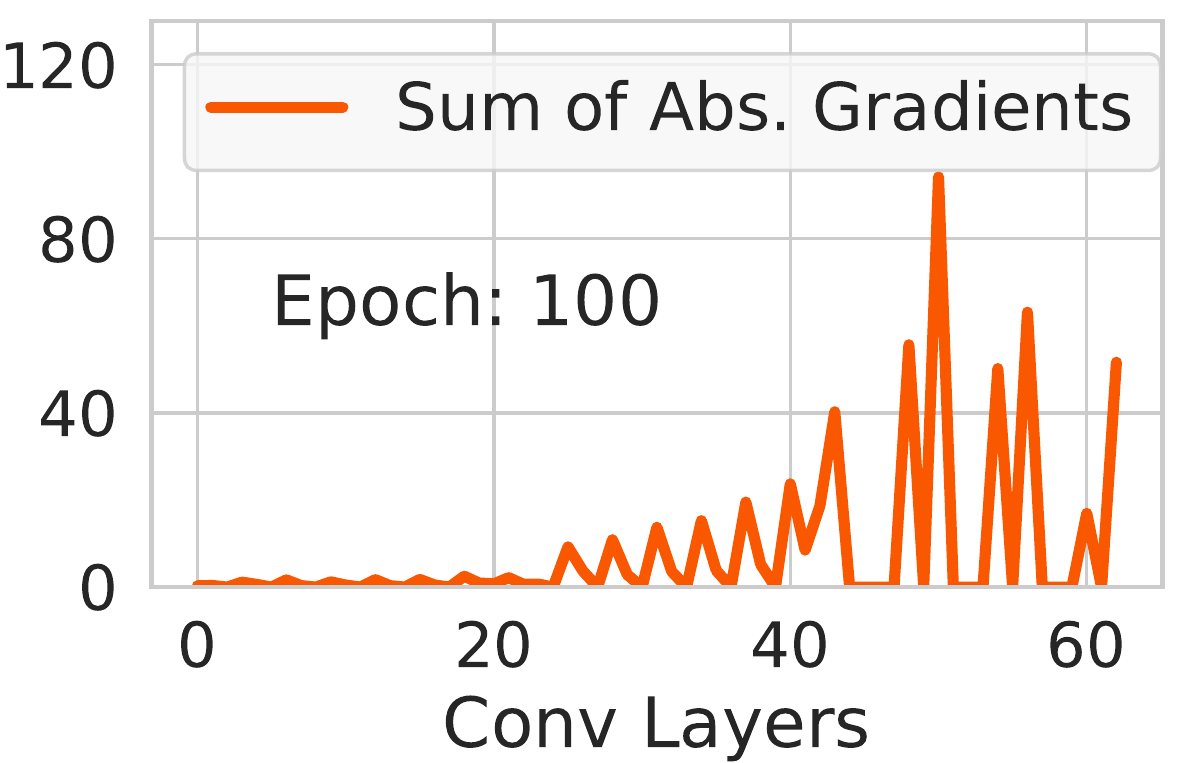}
\includegraphics[width=0.31\columnwidth]{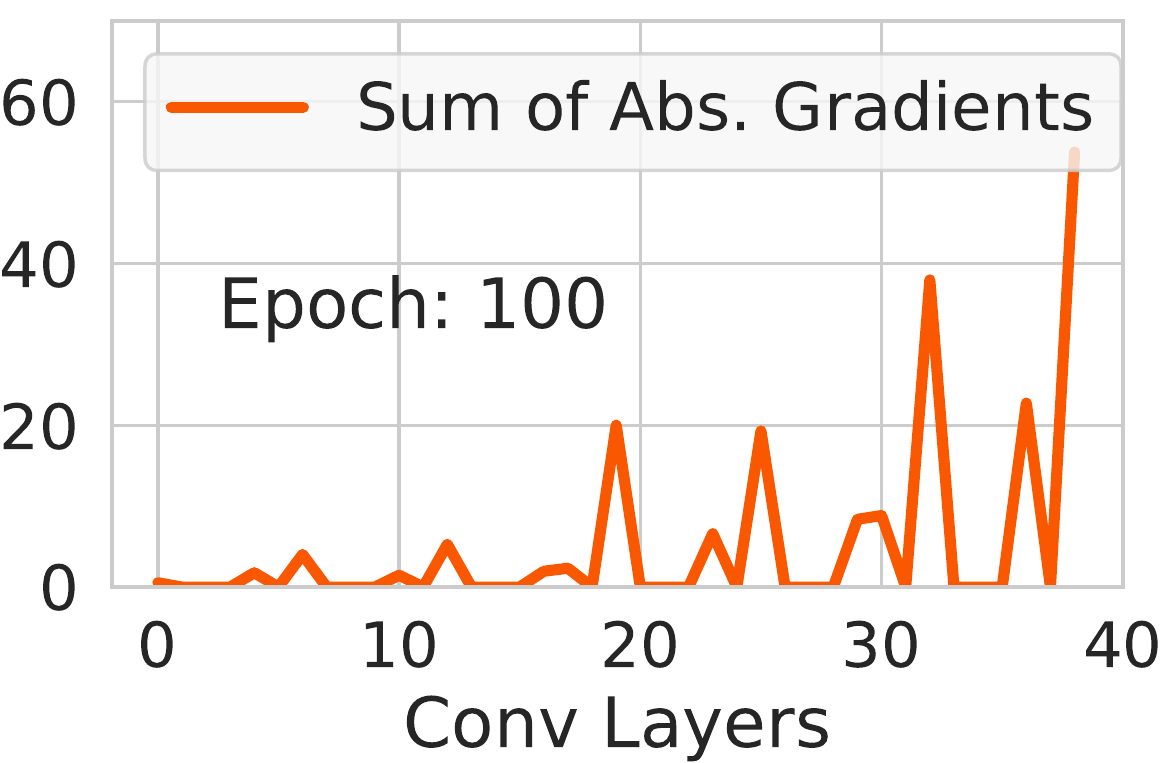}
\includegraphics[width=0.31\columnwidth]{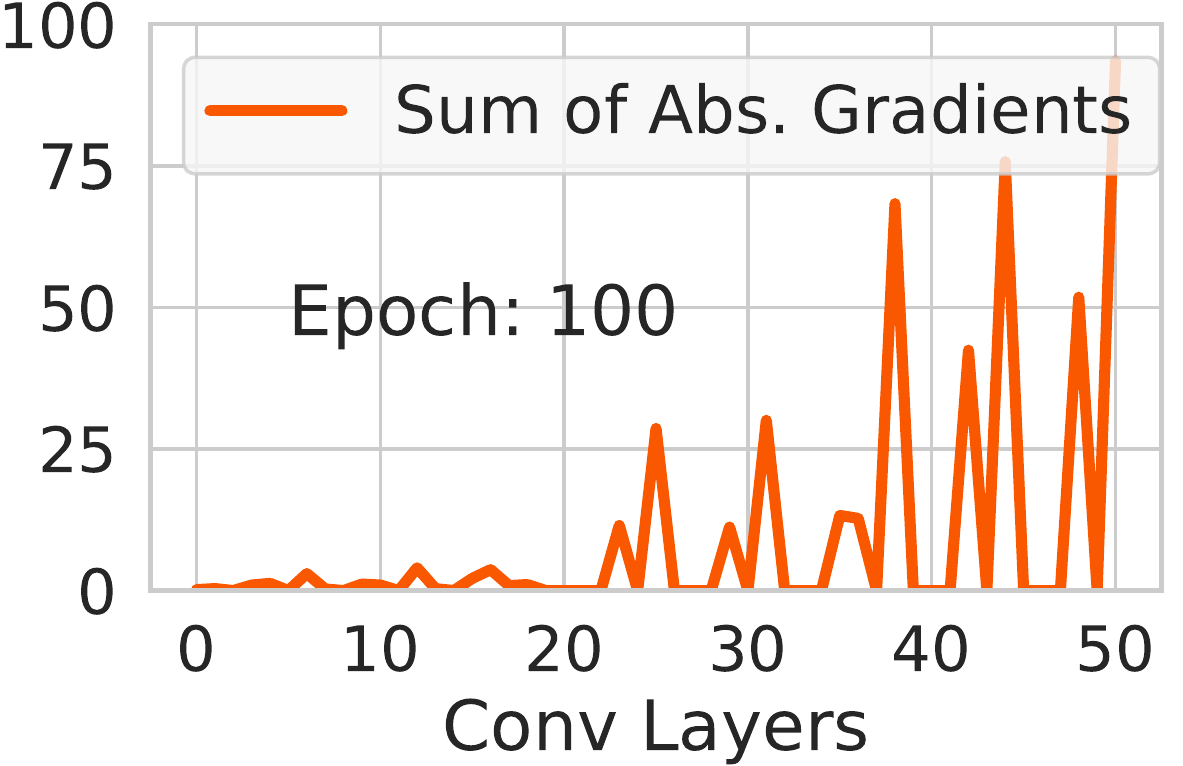}\\
{\small \hspace{1em}(a) BNAS-A w/o SC \hspace{3em}(b) BNAS-B w/o SC \hspace{3em}(c) BNAS-C w/o SC}
\caption{\textbf{Unstable gradients in the `No Skip' ablation (Similar to `w/o SC' in Figure~\ref{fig:unstable_grads}-(b)) of BNAS-\{A,B,C\} models.} We show the sum of gradient magnitudes for convolution layers in all three models for the \textit{No Skip} setting. All three models show spiky gradients without the proposed skip-connections}
\label{fig:no_skip_grads}
\end{figure}

In Figure~\ref{fig:no_skip_grads}, which shows the gradients of the ablated models at epoch 100 similar to Figure~\ref{fig:unstable_grads}, we again observe that the ablated BNAS-\{A,B,C\} without the inter-cell skip connections have unstable (spiky) gradients.
We additionally provide temporally animated plots of the gradients to demonstrate how they change at every 10 epochs starting from 100 epoch to 600 epoch in the accompanied animated gif file (uploaded at \url{https://github.com/gistvision/bnas/blob/master/comb_grads.gif}) -- {\fontfamily{cmss}\selectfont `comb\_grads.gif'}. Table~\ref{table:gif_summ} shows the details of the plots in {\fontfamily{cmss}\selectfont the `comb\_grads.gif'} file.

\begin{table}[h!]
    \centering
    \caption{\textbf{Plot details in {\fontfamily{cmss}\selectfont `comb\_grads.gif'} file.} We provide an animated plot for `BNAS-A w/ SC' for comparison to those of the other ablated models without the skip connection (BNAS-A w/o SC, BNAS-B w/o SC, BNAS-C w/o SC). Other models (BNAS-B and C) with skip connections show similar trend with BNAS-A, and thus omitted for clear presentation}
    \resizebox{0.98\linewidth}{!}{
    \begin{tabular}{ccccc}
    \toprule
    Plot Title  & BNAS-A w/ SC & BNAS-A w/o SC & BNAS-B w/o SC & BNAS-C w/o SC\\ \midrule
    Model   &  BNAS-A & BNAS-A & BNAS-B & BNAS-C \\
    Skip Connections & \color{ForestGreen}\cmark & \color{red}\xmark & \color{red}\xmark & \color{red}\xmark \\
    \bottomrule
    \end{tabular}
    }
    \label{table:gif_summ}
\end{table}

Note that the ablated models (`BNAS-A w/o SC', `BNAS-B w/o SC' and `BNAS-C w/o SC') have unstable (spiky) gradients in the early epochs while the full model (`BNAS-A w/ SC') shows relatively stable (less spiky) gradients in all epochs.
All models eventually show small gradients, indicating the models have stopped learning. 
However, while the training curve of the full model (Figure~\ref{fig:unstable_grads}) implies that it has converged to a reasonable local optima, the training curves of the ablated models (Figure \ref{fig:no_skip_train_curve}) imply that they converged to a poor local optima instead.

\subsection{`No Zeroise' Setting}

In addition to reducing the quantization error, the \textit{Zeroise} layers has additional benefits of more memory savings, reduced FLOPs and more inference speed-up as it does not require any computation and has no learnable parameters.

We summarize the memory savings, FLOPs and inference speed-up of our BNAS-A model in Table~\ref{table:zeroise} by comparing our BNAS-A model with and without \textit{Zeroise} layers.
With the \textit{Zeroise} layers, not only does the accuracy increase, but we also observe significantly more memory savings and inference speed-up.

\label{sec:no_zeroise}
\begin{table}[h!]
\centering
\caption{\textbf{Comparing our searched binary networks (BNAS-A) with and without the Zeroise layer on CIFAR10.} 
\textit{Test. Acc. (\%)} indicates the test accuracy on CIFAR10. The two models compared have the exact same configuration except the usage of \textit{Zeroise} layers in the search space.
Note that the memory savings and inference speed-up were calculated with respect to the floating point version of BNAS-A without \textit{Zeroise} since \textit{Zeroise} layers are not used in floating point domain (see Section~\ref{sec:memory_saving} for related discussion)}
\begin{tabular}{cccc}
\toprule
BNAS-A                    & w/o Zeroise & w/ Zeroise  \\ \midrule
\# Cells/\# Chn. & $20$ / $36$          & $20$ / $36$      \\ 
 Memory Savings   & $31.79\times$   & ${\bf 91.06\times}$ \\ 
 FLOPS ($\times10^8$)      &      $0.36$       &       ${\bf 0.14}$  \\ 
 Inference Speed-up      &      $58.01\times$       &      ${\bf 149.14}\times$  \\ \midrule
Test. Acc.  (\%)     &   $89.47$               & ${\bf 92.70}$         \\ 
\bottomrule
\end{tabular}
\label{table:zeroise}
\end{table}

\section{Additional Ablation Study - No Dilated Convolution Layers}
\label{sec:power_of_search}

\begin{figure}[h!]
\centering
\includegraphics[width=0.30\columnwidth]{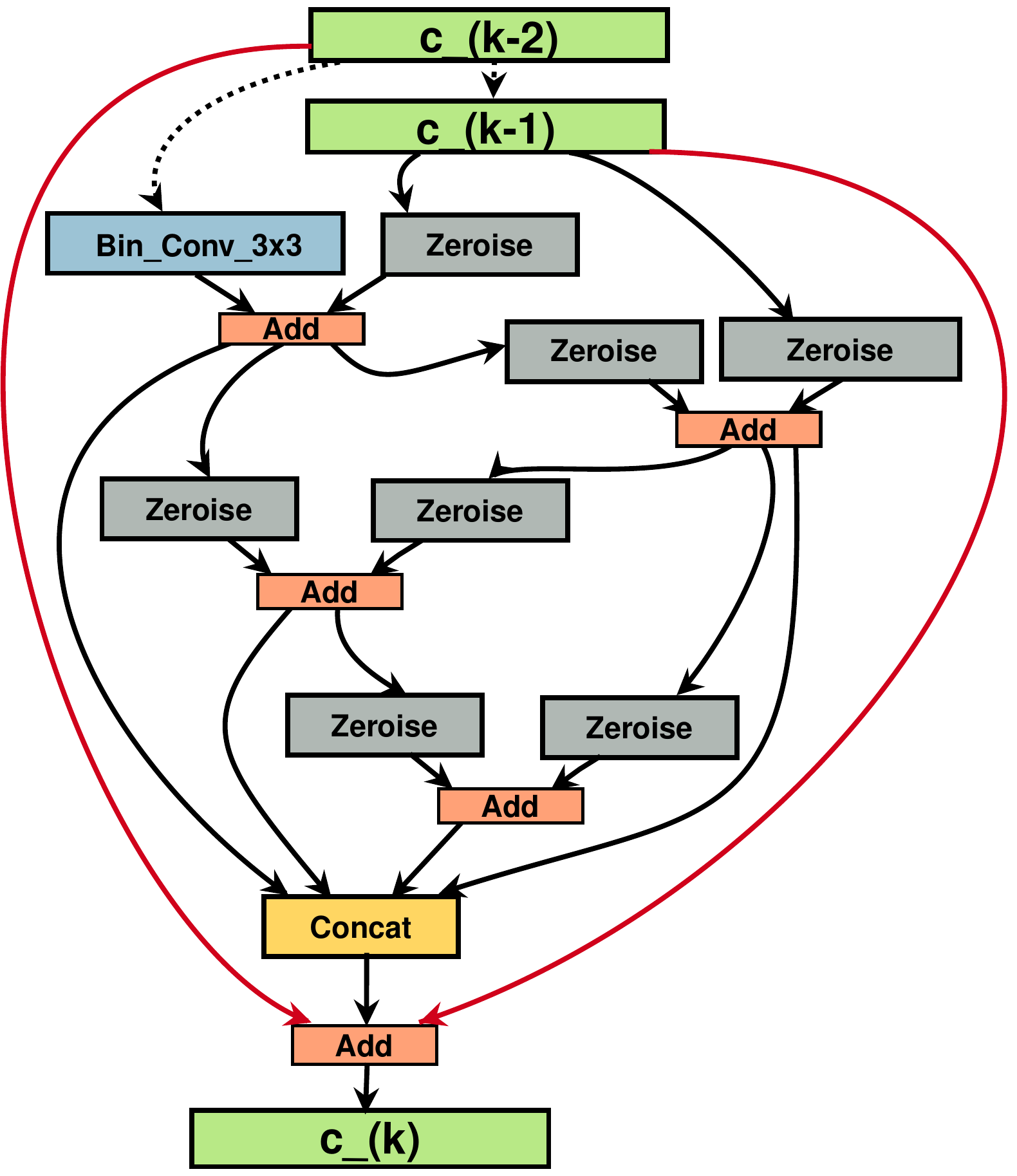}\hspace{4em}
\includegraphics[width=0.30\columnwidth]{images/ours_normal.pdf}\\
\vspace{-.1em}
{\small \hspace{1em}(a) Searched cell w/o binary dil. conv.\hspace{3em}(b) Searched cell w/ binary dil. conv.} 
\vspace{0.2em}
\caption{\textbf{Comparing the cell searched with and without binary dilated convolutions in the search space}. \textit{c\textunderscore(k)} indicates the output of the $k^\text{th}$ cell. The dotted lines represents the connections from the second previous cell (\textit{c\textunderscore(k-2)}). Red lines indicate the inter-cell skip connections}
\label{fig:no_dil_geno}
\end{figure}

To see the impact of dilated convolutions \cite{Yu2015MultiScaleCA} on the performance of our searched architectures, we search binary networks without the dilated convolutions in our search space.
We qualitatively compare the searched cells with and without the dilated convolution layer types in Figure~\ref{fig:no_dil_geno} and quantitatively compare them in Table~\ref{table:no_dil_comp}.
As shown in Table~\ref{table:no_dil_comp}, the dilated convolutional layers contribute to the accuracy of the searched model with a marginal gain.

\begin{table}[h!]
\centering
\caption{\textbf{No dilated convolutions in the search space.} \textit{Test Acc. (\%)} indicates the test accuracy on CIFAR10. \textit{BNAS-A w/o Dil. Conv.} indicates the cell searched without dilated convolutions. The searched architectures perform similarly regardless of dilated convolutions being included or not}

\begin{tabular}{ccc}
\toprule
FLOPs ($\times10^8$)                & Model      & Test Acc. (\%) \\ \midrule
\multirow{2}{*}{$\sim 0.16$} 
    & BNAS-A w/o Dil. Conv.      &   $92.22$      \\ 
    & BNAS-A (w/ Dil. Conv.)     & ${\bf 92.70}$  \\ 
\bottomrule
\end{tabular}
\vspace{-0.5em}
\label{table:no_dil_comp}
\end{table}

\section{Additional Discussions on Separable Convolution} 
\label{sec:bin_nas}

In Section~\ref{sec:binarizing_fpnas}, we claim two issues for the failure of binarized DARTS, SNAS and GDAS; 1) accumulation of quantization error due to separable convolutions, 2) the lack of inter-cell skip connections that makes propagating the gradients across multiple cells difficult.
Particularly, for the first issue (\textit{i.e.}, using separable convolutions accumulates quantization error repetitively), we proposed to exclude the separable convolutions from the search space.
Here, we further investigate the accuracy of the searched architecture with the separable convolutions kept in the search space for binary networks and summarize the results in Table~\ref{table:order_bin}.

Since DARTS, SNAS, and GDAS search on the floating point domain, their search methods do not take quantization error into account and thus result in cells that have a relatively high percentage of separable convolutions and show low test accuracy.
In contrast, we search directly on the binary domain which enables our search method to identify that separable convolutions have high quantization error and hence obtain a cell that contains very few separable convolution (\textit{e.g.}, proportion of separable convolutions is 12.5\% for BNAS while for others, it is higher than 36\%).
Note that explicitly excluding the separable convolutions from the search space does result in better performing binary architectures.
The reason for the failure of separable convolutions is discussed in Section~\ref{sec:sep_conv}.

\begin{table}[h!]
\centering
\caption{\textbf{Effect of separable convolutions in the search space.} \textit{Test Accuracy (\%)} refers to the test accuracy on CIFAR10. \textit{Proportion of Sep. Conv.} refers to the percentage of separable convolutions in the searched convolutional cells. \textit{BNAS-A w/ Sep. Conv.} refers to our searched network with separable convolutions included in the search space. \textit{BNAS-A} refers to our searched network without separable convolutions in the search space. Note that the same binarization scheme (XNOR-Net) was applied in all methods. The models with higher proportion of separable convolutions show lower test accuracy. Please refer to Figure~\ref{fig:search_constraint} for the learning curves of DARTS + Binarized, SNAS + Binarized, GDAS + Binarized, and BNAS-A}

\begin{tabular}{ccc}
    \toprule
    Method   & Test Accuracy (\%) & Proportion of Sep. Conv. (\%) \\
    \midrule
    DARTS + Binarized   & $10.01$            & $62.50$          \\ 
    SNAS + Binarized    & $41.72$          & $36.75$        \\  
    GDAS + Binarized    & $43.19$          & $36.75$         \\ 
    \midrule
    \rowcolor{Gray}
    BNAS-A w/ Sep. Conv.         & $89.66$          & $12.50$ \\
    \rowcolor{Gray}
    BNAS-A &  ${\bf 92.70}$          & $0.00$          \\ 
    \bottomrule
\end{tabular}
\label{table:order_bin}
\end{table}

\section{Additional Discussion on Memory Saving and Inference Speed-up of Our Method}
\label{sec:memory_saving}

Following that other binary networks compare memory savings and inference speed-up with respect to their floating point counterpart~\cite{liu2018bi}, we compute the memory savings and inference speed-up by comparing it to the floating point version of our searched binary networks and summarize the results in Table~\ref{table:comp_savings} for the models for experiments with ImageNet dataset.

Note that all our models achieve higher or comparable memory savings and inference speed-up for the respective FLOPs budgets compared to Bi-Real models \cite{liu2018bi}.

\begin{table}[h!]
    \centering
    \caption{\textbf{Memory savings and inference speed-up compared to floating point counter part of our searched binary network (models for ImageNet experiments).} Note that the memory savings and inference speed-up differ for different networks, as described in~\cite{Rastegari2016XNORNetIC}. The FLOPs, memory savings and inference speed-up for Bi-Real Net is from Table 3 in \cite{liu2018bi}. All our models achieve higher or comparable memory savings and inference speed-up compared to Bi-Real Net~\cite{liu2018bi}}
    \resizebox{0.988\linewidth}{!}{
    \begin{tabular}{cccccccc}
    \toprule
    Model     & BNAS-D & BNAS-E & BNAS-F & BNAS-G & BNAS-H & Bi-Real (Bi-Real Net18)& Bi-Real (Bi-Real Net34)\\ \cmidrule(lr){1-1} \cmidrule(lr){2-6}\cmidrule(lr){7-8}
    FLOPs ($\times10^{8}$) & $\sim1.48$ & $\sim 1.63$  & $\sim 1.78$ & $\sim1.93$ & $\sim 6.56$ & $\sim 1.63$ & $\sim1.93$\\ 
    Memory Savings  & $13.91\times$ & $14.51\times$  & $30.37\times$  &  $14.85\times$ &  $21.75\times$&$11.14\times$& $15.97\times$\\
    Inference Speed-up  & $20.85\times$ & $21.15\times$  & $24.29\times$  &  $19.34\times$& $24.81\times$ & $11.06\times$& $18.99\times$\\
    \bottomrule
    \end{tabular}
    }
    \label{table:comp_savings}
\end{table}

\section{Additional Remarks on Quantized (`Non 1-bit') or not fully binary CNNs}
\label{sec:quantized_cnn}

In Section~\ref{sec:intro}, we mention that binary networks or 1-bit CNNs are distinguished from quantized networks (using more than 1 bit) and not fully binary networks (networks only with binary weights but floating point activations) due to the extreme memory savings and inference speed-up they bring.
Quantized or not fully binarized networks that incorporate search are a type of efficient networks that are not comparable to 1-bit CNNs because they cannot utilize XNOR and bit counting operations in the inference which significantly brings down their memory savings and inference speed up gains.

It is, however,  interesting to note that there are a line of work for efficient networks with more resource consumption, especially the recent ones.
Notably, \cite{Chen2018JointNA, wang2019haq, Wu2018MixedPQ, Lou2019AutoQBAF} search for multi-bit quantization policies only and solely \cite{Chen2018JointNA} search for network architectures as well.
\cite{Chen2019BinarizedNA} also search for network architectures for binary weight (not fully binarized) CNNs.
Their networks are not fully binarized (networks only with binary weights) which makes them incomparable to other binary networks.
Moreover, \cite{wang2019haq, Wu2018MixedPQ, Lou2019AutoQBAF} all search for quantization policies, not network architectures, further differentiating it from our method.

\end{document}